\PassOptionsToPackage{table}{xcolor}
\PassOptionsToPackage{sort&compress}{natbib}
\documentclass[onecolumn]{dataflow}
\usepackage{amsmath,amssymb,tabularx,adjustbox}
\newif\ifCodeVBTableFloat
\AtBeginEnvironment{table}{\CodeVBTableFloattrue}
\AtBeginEnvironment{table*}{\CodeVBTableFloattrue}
\let\CodeVBNativeTabular\tabular
\let\CodeVBNativeEndTabular\endtabular
\expandafter\let\expandafter\CodeVBNativeTabularStar\csname tabular*\endcsname
\expandafter\let\expandafter\CodeVBNativeEndTabularStar\csname endtabular*\endcsname
\newsavebox{\CodeVBZeroTableBox}
\newsavebox{\CodeVBUnitTableBox}
\newsavebox{\CodeVBFullWidthTableBox}

\ExplSyntaxOn
\tl_new:N \l__codevb_columns_tl
\dim_new:N \l__codevb_zero_dim
\dim_new:N \l__codevb_step_dim
\cs_new_protected:Npn \__codevb_native_table:nnn #1#2#3
  {\CodeVBNativeTabular[#1]{#2}#3\CodeVBNativeEndTabular}
\cs_new_protected:Npn \__codevb_fixed_table:nnn #1#2#3
  {
    \group_begin:
      \cs_set_eq:NN \CodeVBSavedMulticolumn \multicolumn
      \cs_set:Npn \multicolumn ##1##2##3
        {\CodeVBSavedMulticolumn{##1}{##2}{}}
      \dim_zero:N \tabcolsep
      \sbox{\CodeVBZeroTableBox}{\__codevb_native_table:nnn {#1}{#2}{#3}}
      \dim_set:Nn \tabcolsep {1pt}
      \sbox{\CodeVBUnitTableBox}{\__codevb_native_table:nnn {#1}{#2}{#3}}
      \cs_set_eq:NN \multicolumn \CodeVBSavedMulticolumn
      \dim_set:Nn \l__codevb_zero_dim {\box_wd:N \CodeVBZeroTableBox}
      \dim_set:Nn \l__codevb_step_dim
        {\box_wd:N \CodeVBUnitTableBox-\l__codevb_zero_dim}
      \dim_set:Nn \tabcolsep
        {\fp_to_dim:n {max(0,
          (\dim_to_fp:n{\textwidth}-\dim_to_fp:n{\l__codevb_zero_dim})
          /\dim_to_fp:n{\l__codevb_step_dim})}}
      \sbox{\CodeVBFullWidthTableBox}{\__codevb_native_table:nnn {#1}{#2}{#3}}
      \typeout{CodeVB~table~\thetable\c_colon_str~natural~width=\the\wd\CodeVBFullWidthTableBox;~target=\the\textwidth;~padding=\the\tabcolsep}
      \begin{adjustbox}{max~width=\textwidth}
        \clipbox{0pt~0pt~0pt~0pt}{\usebox{\CodeVBFullWidthTableBox}}
      \end{adjustbox}
    \group_end:
  }
\NewDocumentCommand{\CodeVBFullWidthTable}{m m +m}
  {\__codevb_fixed_table:nnn {#1}{#2}{#3}}
\NewDocumentCommand{\CodeVBFullWidthStarTable}{m m +m}
  {
    \tl_clear:N \l__codevb_columns_tl
    \clist_map_inline:nn
      {lr,lrr,lrrr,lrrrr,lrrrrr,lrrrrrr,lrrrrrrr,lrrrrrrrr,lrrrrrrrrr,lrrrrrrrrrr}
      {
        \tl_if_eq:nnT {#2} {@{\extracolsep{\fill}}##1@{}}
          {\tl_set:Nn \l__codevb_columns_tl {##1}}
      }
    \tl_if_eq:NnT \l__codevb_columns_tl {lrrrrrrr}
      {
        \int_compare:nNnT {\value{table}} = {1}
          {\tl_set:Nn \l__codevb_columns_tl
            {>{\raggedright\arraybackslash}p{0.28\textwidth}rrrrrrr}}
      }
    \tl_if_empty:NTF \l__codevb_columns_tl
      {\PackageError{CodeVB-tables}{Unexpected~tabular*~column~specification}{}}
      {\exp_args:NnV \__codevb_fixed_table:nnn {#1}\l__codevb_columns_tl {#3}}
  }
\ExplSyntaxOff

\RenewDocumentEnvironment{tabular}{O{c} m +b}{%
  \ifCodeVBTableFloat
    \CodeVBFullWidthTable{#1}{#2}{#3}%
  \else
    \begin{adjustbox}{max width=\linewidth}%
      \CodeVBNativeTabular[#1]{#2}#3\CodeVBNativeEndTabular
    \end{adjustbox}%
  \fi
}{}
\RenewDocumentEnvironment{tabular*}{m O{c} m +b}{%
  \ifCodeVBTableFloat
    \CodeVBFullWidthStarTable{#2}{#3}{#4}%
  \else
    \begin{adjustbox}{max width=\linewidth}%
      \CodeVBNativeTabularStar{#1}[#2]{#3}#4\CodeVBNativeEndTabularStar
    \end{adjustbox}%
  \fi
}{}
\AtBeginEnvironment{tabularx}{%
  \let\tabular\CodeVBNativeTabular
  \let\endtabular\CodeVBNativeEndTabular
  \expandafter\let\csname tabular*\endcsname\CodeVBNativeTabularStar
  \expandafter\let\csname endtabular*\endcsname\CodeVBNativeEndTabularStar
}

\newif\ifCodeVBSingleColumnFigure
\AtBeginEnvironment{figure}{%
  \CodeVBSingleColumnFiguretrue
  \captionsetup{width=\textwidth}%
}
\AtBeginEnvironment{figure*}{%
  \CodeVBSingleColumnFigurefalse
  \captionsetup{width=\textwidth}%
}
\let\CodeVBUnscaledIncludeGraphics\includegraphics

\NewDocumentCommand{\CodeVBDrawOriginalGraphics}{m m m}{%
  \IfBooleanTF{#1}{%
    \CodeVBUnscaledIncludeGraphics*[#2]{#3}%
  }{%
    \CodeVBUnscaledIncludeGraphics[#2]{#3}%
  }%
}

\RenewDocumentCommand{\includegraphics}{s O{} m}{%
  \ifCodeVBSingleColumnFigure
    \ifdim\linewidth=\textwidth
      \typeout{CodeVB single-column graphic: #3; width=\the\textwidth}%
      \ifstrempty{#2}{%
        \CodeVBDrawOriginalGraphics{#1}{width=\textwidth,height=0.80\textheight,keepaspectratio}{#3}%
      }{%
        \CodeVBDrawOriginalGraphics{#1}{#2,width=\textwidth,height=0.80\textheight,keepaspectratio}{#3}%
      }%
    \else
      \CodeVBDrawOriginalGraphics{#1}{#2}{#3}%
    \fi
  \else
    \CodeVBDrawOriginalGraphics{#1}{#2}{#3}%
  \fi
}

\let\CodeVBFloatFigure\figure
\let\CodeVBFloatEndFigure\endfigure
\expandafter\let\expandafter\CodeVBFloatWideFigure\csname figure*\endcsname
\expandafter\let\expandafter\CodeVBFloatEndWideFigure\csname endfigure*\endcsname
\RenewDocumentEnvironment{figure}{O{!htbp}}{%
  \CodeVBFloatFigure[!htbp]%
}{\CodeVBFloatEndFigure}
\RenewDocumentEnvironment{figure*}{O{!htbp}}{%
  \CodeVBFloatWideFigure[!htbp]%
}{\CodeVBFloatEndWideFigure}

\let\CodeVBFloatTable\table
\let\CodeVBFloatEndTable\endtable
\expandafter\let\expandafter\CodeVBFloatWideTable\csname table*\endcsname
\expandafter\let\expandafter\CodeVBFloatEndWideTable\csname endtable*\endcsname
\RenewDocumentEnvironment{table}{O{!htbp}}{%
  \CodeVBFloatTable[!htbp]%
}{\CodeVBFloatEndTable}
\RenewDocumentEnvironment{table*}{O{!htbp}}{%
  \CodeVBFloatWideTable[!htbp]%
}{\CodeVBFloatEndWideTable}

\AtBeginDocument{%
  \pretocmd{\section}{\ifnum\value{section}>0 \FloatBarrier\fi}{}{%
    \PackageError{CodeVB-layout}{Cannot attach section float barrier}{}%
  }%
  \pretocmd{\subsection}{\FloatBarrier}{}{%
    \PackageError{CodeVB-layout}{Cannot attach subsection float barrier}{}%
  }%
}

\renewcommand{\templatelogo}{%
  \includegraphics[height=9mm]{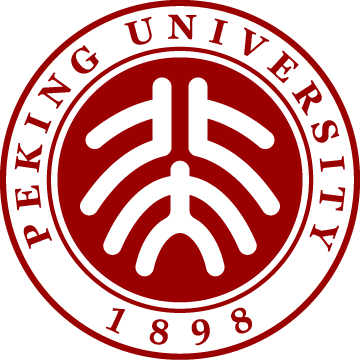}%
  \hspace{3mm}%
  \includegraphics[height=9mm]{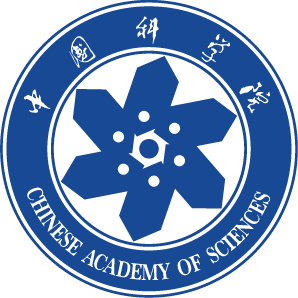}%
  \hspace{3mm}%
  \includegraphics[height=9mm]{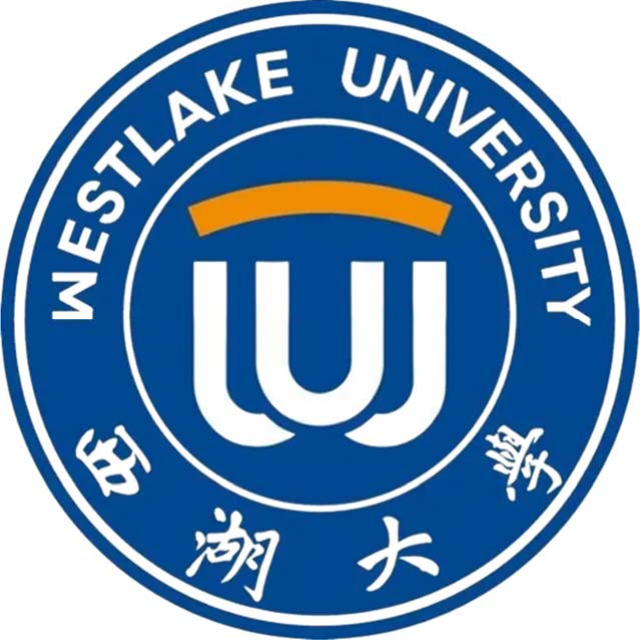}%
  \hspace{4mm}%
  \includegraphics[height=9mm]{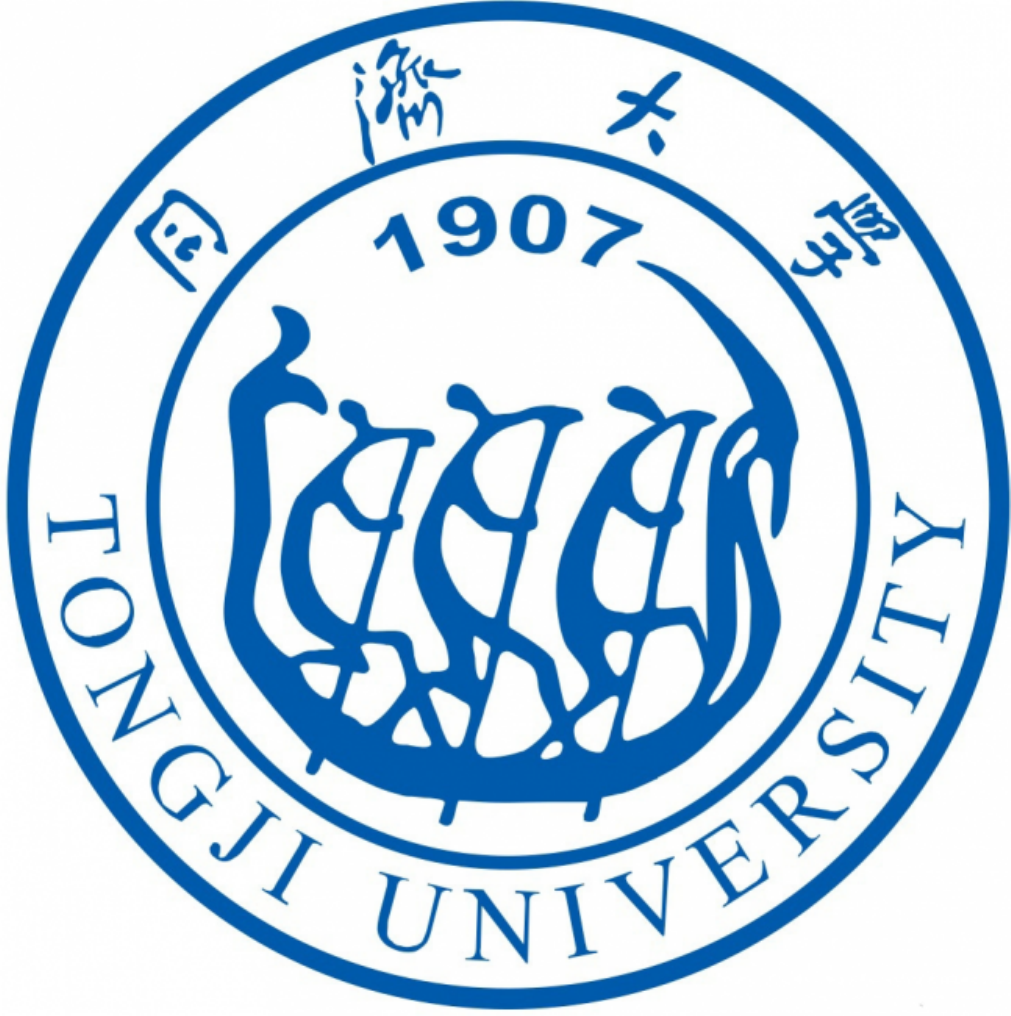}%
  \hspace{4mm}%
  \includegraphics[height=9mm]{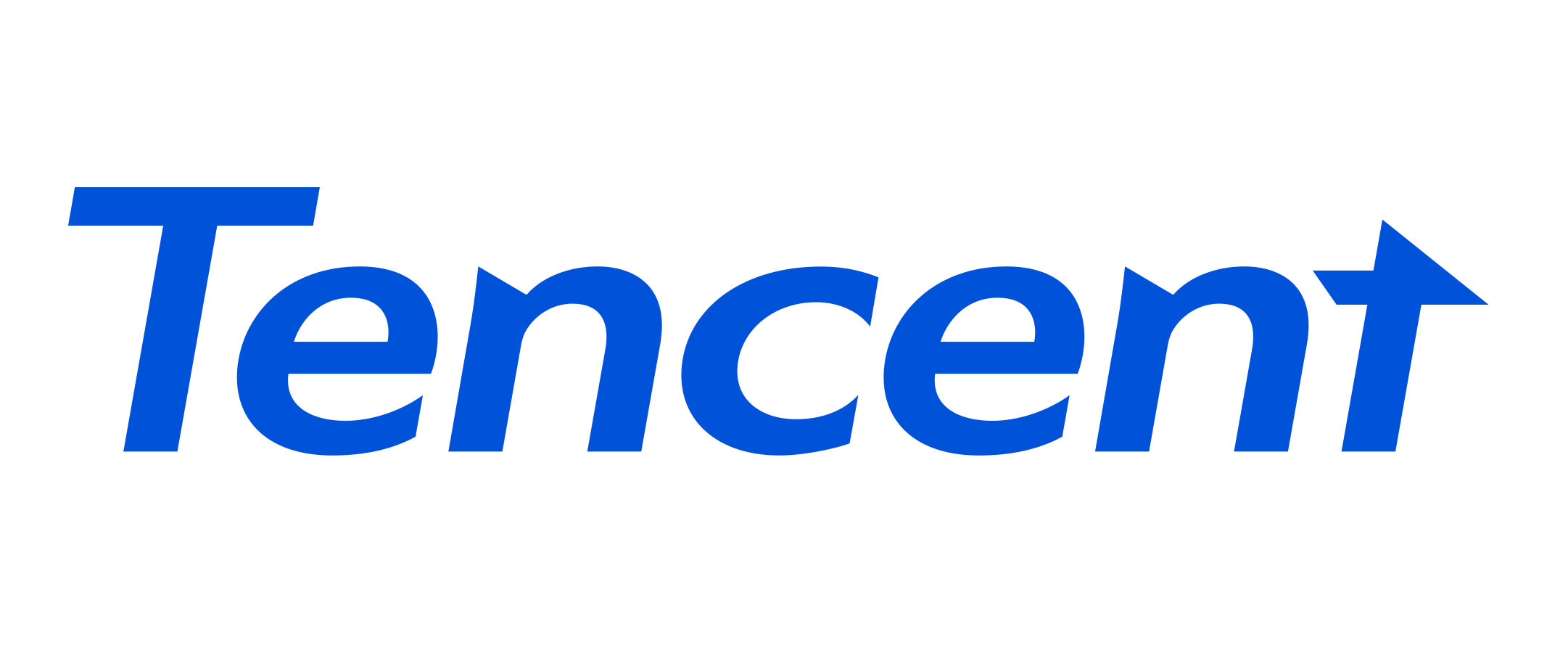}%

}
\fancypagestyle{firststyle}{%
  \fancyhf{}
  \renewcommand{\headrulewidth}{0pt}
  \fancyhead[L]{\templatelogo}
  \fancyhead[R]{%
    \raisebox{2.5mm}{\normalfont\small\textcolor[HTML]{4D4D4D}{\today}}%
  }
}
\renewcommand{\titlefont}{%
  \sffamily\fontsize{16}{20}\selectfont\raggedright
  \hyphenpenalty=10000\relax\exhyphenpenalty=10000\relax
}
\DeclareFontShape{T1}{bytesans}{m}{it}{<->ssub*lmss/m/sl}{}
\DeclareFontShape{T1}{bytesans}{b}{it}{<->ssub*lmss/bx/sl}{}
\DeclareFontShape{T1}{bytesans}{bx}{it}{<->ssub*lmss/bx/sl}{}
\setcitestyle{numbers,square,comma}
\providecommand{\thetitle}{{\sffamily\fontsize{16}{20}\selectfont\thetitletext}}

\let\CodeVBOriginalAppendix\appendix
\renewcommand{\appendix}{%
  \CodeVBOriginalAppendix
  \renewcommand{\thetitle}{{\sffamily\fontsize{16}{20}\selectfont\thetitletext}}%
  \let\CodeVBAppendixSetCounter\setcounter
  \renewcommand{\setcounter}[2]{%
    \ifstrequal{##1}{page}{}{\CodeVBAppendixSetCounter{##1}{##2}}%
  }%
  \renewcommand{\pagenumbering}[1]{}%
}

\let\CodeVBTemplateAbstract\abstract
\let\CodeVBTemplateMakeTitle\maketitle
\renewcommand{\maketitle}{}
\RenewDocumentEnvironment{abstract}{+b}{%
  \CodeVBTemplateAbstract{#1}%
  \CodeVBTemplateMakeTitle
}{}

\RenewDocumentCommand{\addtolist}{O{}+m+m+m+m}{%
  \ifdefempty{#3}{}{\gappto#3{#5}}%
  \gappto#3{#4[#1]{#2}}%
}

\renewcommand{\email}[1]{%
  \href{mailto:#1}{\textcolor{black}{\normalfont\small\detokenize{#1}}}%
}

\renewcommand{\checkdataformat}[2][]{%
  \par\noindent
  \parbox[t]{64pt}{%
    \checkdatafont\sffamily\bfseries\strut\mbox{#1:}}%
  \parbox[t]{\dimexpr\linewidth-64pt\relax}{%
    \normalfont\small\raggedright\strut #2}\par
}

\makeatletter
\renewcommand{\mymaketitle}{%
  \vspace*{-4mm}%
  \@toptitlebar
  \begin{tcolorbox}[
    enhanced,colback=boxbg,frame hidden,arc=3mm,
    left=8mm,right=8mm,top=5mm,bottom=5mm,
    before skip=0pt,after skip=0pt
  ]
    \setlength{\parskip}{0pt}
    \setlength{\parindent}{0pt}
    {\titlelist\par}
    \ifdefempty{\authorlist}{}{\vskip 3mm {\centering\authorlist\par}}
    \ifdefempty{\affiliationlist}{}{\vskip 2mm {\centering\affiliationlist\par}}
    \vskip 3mm
    \abstractlist\par
    \ifdefempty{\contributionlist}{}{%
      \vskip 2mm
      \contributionlist\par
    }
    \ifdefempty{\checkdatalist}{}{%
      \vskip 2mm
      \checkdatalist\par
    }
  \end{tcolorbox}
  \tcbset{reset}
  \FloatBarrier
  \thispagestyle{firststyle}
}
\makeatother
\renewcommand{\CodeVBTemplateMakeTitle}{\mymaketitle\vskip 4mm}

\usepackage{cuted} 

\usepackage{currfile} 

\usepackage{caption} 
\definecolor{cvprblue}{rgb}{0.21,0.49,0.74}

\usepackage{booktabs}
\usepackage{multirow}
\usepackage{amsmath}
\usepackage{tabularx}
\usepackage{placeins}

\usepackage[table]{xcolor}

\definecolor{bestblue}{RGB}{226,238,249}

\usepackage{tikz}
\newsavebox{\caseimage}
\usepackage{graphicx}
\usepackage{xcolor}
\usepackage{tikz}

\usepackage[table]{xcolor}

\definecolor{bestblue}{RGB}{226,238,249}

\usepackage{tcolorbox}
\tcbuselibrary{skins,breakable}
\usepackage{enumitem}

\newtcolorbox{promptcard}[2][]{
  enhanced,
  breakable,
  colback=white,
  colframe=black!22,
  colbacktitle=orange!7,
  coltitle=black,
  title={#2},
  fonttitle=\bfseries\small,
  fontupper=\footnotesize,
  fontlower=\footnotesize,
  boxrule=0.45pt,
  arc=0.8mm,
  left=1.8mm,
  right=1.8mm,
  top=1.4mm,
  bottom=1.4mm,
  titlerule=0.35pt,
  borderline west={1.8pt}{0pt}{orange!65!black},
  before skip=7pt,
  after skip=9pt,
  #1
}

\title{CodingVBench: Rethinking Video Quality Evaluation Beyond Generative Models}

\author[1,*]{Yuxia Zhu}
\author[2,*]{Junhan Zhu}
\author[3,*]{Shiying Zhang}
\author[4]{Xinglong Xu}
\author[4]{Linzhuang Sun}
\author[1]{Hao Liang}
\author[4]{Caijun Jia}
\author[1]{Qifeng Cai}
\author[1]{Jincen Luo}
\author[5]{Alden Cai}
\author[5,\dagger]{Asher Qin}
\author[1,\dagger]{Wentao Zhang}
\author[5,\dagger]{Cheng Tan}

\affiliation[]{%
  $^{1}$Peking University\quad
  $^{2}$Westlake University \quad
  $^{3}$Tongji University\\
  $^{4}$University of Chinese Academy of Sciences\quad
  $^{5}$Tencent
}
\contribution[*]{Equal contribution.}
\contribution[\dagger]{Corresponding author.}

\newcommand{\contactlabel}[2]{%
  \makebox[11pt][c]{#1}\hspace{3pt}#2}
\newcommand{\emailicon}{\raisebox{-1.5pt}{\includegraphics[height=1.05em]{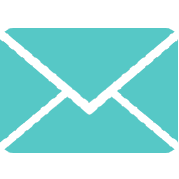}}}
\newcommand{\sourcelink}{\raisebox{-1.5pt}{\includegraphics[height=1.05em]{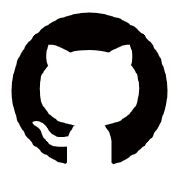}}}

\newcommand{\websitelink}{\raisebox{-1.5pt}{\includegraphics[height=1.05em]{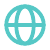}}}

\checkdata[\contactlabel{\emailicon}{Email}]{%
  \email{Zzzyxiia@outlook.com}}
\checkdata[\contactlabel{\sourcelink}{Code}]{%
  \href{https://github.com/OpenRaiser/CodingVBench}{https://github.com/OpenRaiser/CodingVBench}}
\checkdata[\contactlabel{\websitelink}{Website}]{%
  \href{https://openraiser.github.io/CodingVBench}{https://openraiser.github.io/CodingVBench}}

\makeatletter
\AtBeginDocument{%
  \if@twocolumn
    \PackageError{CodeVB-layout}{Expected single-column layout}{Compile main.tex.}%
  \fi
  \typeout{CodeVB layout: single column; columnwidth=\the\columnwidth; textwidth=\the\textwidth}%
}
\makeatother

\begin{document}
\maketitle

\begin{abstract}
    Video generation benchmarks such as VBench assess visual quality and prompt adherence using metrics validated against human judgments on pixel-generated clips. Coding agents now answer the same prompts through executable scene programs. Yet action scores can reverse visible completion: a code-rendered baby wakes and stretches but scores zero, while a pixel-generated baby stays asleep but scores one. Recognition-based semantic scores can accept familiar content without verifying the requested event. We introduce CodingVBench, a benchmark for testing video evaluators across both production routes. It contains 360 matched code-rendered and MiniMax-H3 videos labeled by three annotators for task completion and visual preference, and 60 controlled pairs in which an edit to a verified scene program either removes a required behavior or only shifts the hue. Native modules often give an edited video the same score as its original, accept videos that lack the requested event, and rank aesthetics against human preference. Two eight-frame VLM judges outperform them in agreement with human completion, reaching 92.7\% and 93.9\% against 71.6\%, and the primary judge ranks every behavioral violation below its original. Cross-route comparison therefore needs task completion, visual quality, and scoring coverage reported separately, and we will release the videos, labels, controlled pairs, and evaluation code.
\end{abstract}    

\begin{figure*}[ht]
\centering
\includegraphics[width=\textwidth]{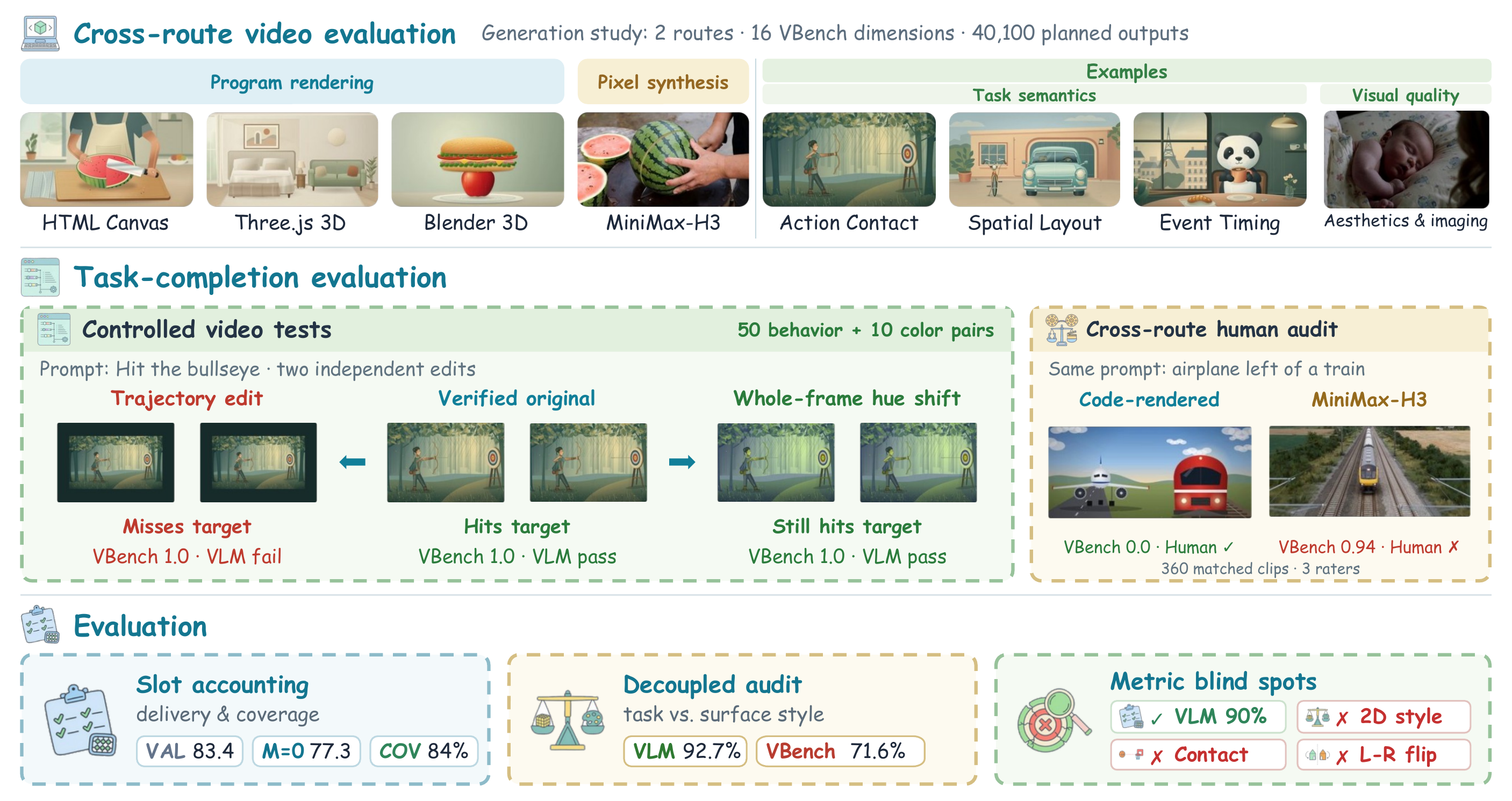}
\caption{\textbf{The same request can be completed by code rendering or pixel synthesis, and the two routes need different checks.} Top: examples separate program rendering from pixel synthesis, and task semantics from visual quality. Middle: trajectory and hue edits test the required hit; three annotators assess 360 matched code and MiniMax-H3 videos. Displayed VLM results are for GPT-6.1 Sol. Bottom: human agreement, paired outcomes, and coverage. VAL and M$=0$ show Astra totals with undelivered videos excluded or assigned zero; COV shows Opus 5.5 color-scoring coverage. The broader generation study contains 40,100 planned output slots.}
\label{fig:overview}
\end{figure*}
\section{Introduction}
\label{sec:intro}

Coding agents now make videos by writing a scene program that a browser or graphics engine renders~\cite{chen2025code2video,liu2025logomotion,chen2026code2video,vibevideo}, alongside models that synthesize pixels directly~\cite{sora,kong2024hunyuanvideo,wan2025wan,yang2025cogvideox}. Both routes answer the same prompts, and VBench and VBench-2.0 score them with metrics whose agreement with human judgments was established on pixel-generated clips~\cite{huang2024vbench,zheng2025vbench}. For ``baby waking up,'' the code-rendered baby wakes and stretches, while the MiniMax-H3 baby stays asleep (Figure~\ref{fig:actions}). VBench gives the first video an action score of zero and the second a one. The video that completes the request scores below the video that fails it.

The main reason is what the scores look for. The action score asks whether a named action is recognizable in the frames, so a scene that resembles the action can pass without the event, and a geometric animation of the event may not resemble the photographs the recognizers were built on. In ``cutting watermelon,'' by contrast, both videos execute the cut and both score one, although one is a photograph and the other is made of primitives. Task completion and visual quality are different questions, and a single score answers neither reliably across routes.

Independently produced videos differ in texture, framing, object appearance, and motion, so an evaluator's failure has no single cause. A scene program lets us change one thing at a time. Redirecting an arrow removes the requested hit while the motion stays continuous, and shifting the hue changes the color while the hit stays visible. A judge that rejects the first video and accepts the second responds to the requested behavior and ignores the color.

We present CodingVBench to test video evaluators across both routes (Figure~\ref{fig:overview}). It contains 360 matched videos for 120 VBench prompts, produced by two coding models and MiniMax-H3, with completion and pairwise visual-preference labels from three annotators. It also contains 60 controlled pairs: 50 remove a required behavior such as contact, motion, a spatial relation, or the order of events, and ten shift the hue while the task stays complete. We compare native semantic modules and two frozen multi-frame VLM judges against completion labels, native aesthetic rankings against preferences, and all evaluators on the controlled pairs.

Code2Video-Compass~\cite{chen2026code2video} and VibeVideoBench~\cite{vibevideo} evaluate code-driven video creation, whereas CodingVBench tests the evaluators of those videos. Unlike PROWBench, which checks rendered events against executable world records~\cite{huang2026rowbench}, and IsoVRBench, which isolates video properties to diagnose reward models~\cite{li2026isovrbench}, we pair human completion labels from both routes with edits that either break a requested behavior or change only color.

Native scores fail in both directions: overall agreement with human completion is 71.6\%, and two coding models with similar human completion rates receive native acceptance of 87.5\% and 59.2\%. Native aesthetic rankings often favor the video that humans rate lower. Eight-frame GPT-6.1 Sol and Claude Sonnet 5.5 raise agreement to 92.7\% and 93.9\%. On the controlled pairs, Sol ranks all 50 violations below their completed originals, including cases that leave native scores unchanged, and accepts both videos in 90\% of the hue pairs. A judge should therefore be chosen by how it responds to edits as well as by its agreement with humans, and cross-route comparisons should report completion, visual quality, and scoring coverage separately.

\section{Related Work}
\label{sec:related}

\subsection{Video Generation Evaluation}
Distributional metrics such as FVD~\cite{unterthiner2018towards} and image--text compatibility scores based on CLIP~\cite{radford2021learning,hessel2021clipscore} summarize visual similarity and alignment. EvalCrafter~\cite{liu2024evalcrafter}, FETV~\cite{liu2023fetv}, and VBench~\cite{huang2024vbench} organize quality and adherence into multiple dimensions. Targeted benchmarks examine composition~\cite{sun2025t2v,li2024genai}, temporal change~\cite{feng2025tc,yuan2024chronomagic}, and dynamics~\cite{liao2024evaluation}; VideoPhy, VideoPhy-2, and VBench-2.0 further test physical commonsense and intrinsic faithfulness~\cite{bansal2025videophy,bansal2026videophy,zheng2025vbench}.

Fine-grained image evaluation provides related ways to check visible requirements. GenEval tests objects, counts, colors, and positions~\cite{ghosh2023geneval}; TIFA and Davidsonian Scene Graph decompose prompts into visual questions~\cite{hu2023tifa,cho2024davidsonian}; VQAScore assesses text--visual alignment through question answering~\cite{lin2024evaluating}. Native video metrics instead combine recognition modules~\cite{umt,grit,tag2text}, video--text and flow models~\cite{internvid,raft}, and perceptual predictors~\cite{dover,musiq}. We test agreement with human completion and visual preference across code rendering and pixel generation, where validation on the original video population may not transfer.

\subsection{Code-Based Video Creation}
Layout and scene systems express spatial structure through explicit layouts, constraints, or programs~\cite{feng2023layoutgpt,yang2024holodeck,hu2024scenecraft,tam2024scenemotifcoder,huang2024blenderalchemy}. Related visual creation systems produce sketches~\cite{text2vector}, animate drawings~\cite{livesketch}, generate HTML/CSS animations~\cite{liu2025logomotion,keyframer}, and render educational scenes~\cite{chen2025code2video}. GPT4Motion and VideoCoCo combine scripted 3D structure with diffusion rendering~\cite{lv2024gpt4motion,videococo}. VideoWeaver and Timeline-Bench evaluate agents for video production and editing~\cite{videoweaver,timeline}, while Code2Video-Compass and VibeVideoBench assess general-purpose code-driven creation~\cite{chen2026code2video,vibevideo}. We use the resulting videos and editable programs to examine evaluator reliability, rather than treating the programs only as outputs for ranking creation capability.

\subsection{Program-Based Evaluator Diagnosis}
Synthetic video datasets isolate compositional actions and causal events~\cite{girdhar2019cater,yi2019clevrer}, with Kubric providing reproducible scene generation~\cite{greff2022kubric}. Winoground, ARO, and SugarCrepe expose weak relational understanding and sensitivity to evaluation shortcuts~\cite{thrush2022winoground,yuksekgonul2022and,hsieh2023sugarcrepe}. TempCompass and Vinoground extend diagnosis to temporal video understanding~\cite{tempcompass,vinoground}. Preference and video-quality models learn from human feedback~\cite{xu2023imagereward,he2024videoscore,he2025videoscore2,tong2025mj,liu2026improving}, but preference agreement and event-level discrimination require different tests. Closest to our setting, PROWBench checks rendered events against executable world records~\cite{huang2026rowbench}, while IsoVRBench diagnoses and trains reward models through isolated video interventions~\cite{li2026isovrbench}. CodingVBench combines cross-route human completion labels with task-specific program edits that remove required behavior or preserve completion under color changes. We verify the rendered pairs to check that the intended difference is visible.

\section{Evaluation Targets}
\label{sec:preliminaries}

We compare videos made for the same text request $p$. A coding agent writes a scene program $z$ that renders a video $v$, while a pixel generator produces $v$ directly. Both videos are judged against the same request.

\paragraph{Task completion.}
Let
\begin{equation}
Q(p)=\{q_1,\ldots,q_K\}
\label{eq:cvb_requirements}
\end{equation}
contain the visible requirements of $p$, including objects, attributes, actions, spatial relations, and temporal events. For ``hit the bullseye,'' an archer and a target are not enough: the arrow must visibly reach the bullseye. We assign $y=1$ when every requirement is visibly satisfied and $y=0$ when one is violated. Cases that cannot be verified because of occlusion or framing remain unresolved.

\paragraph{Visual quality.}
Visual preference is recorded separately. Annotators say which of two videos for the same prompt has the better composition, lighting, texture, and rendering detail, or whether they are comparable. A style or a realistic motion that the prompt asks for belongs to $Q(p)$. When the prompt does not ask for it, a geometric animation can complete the task without photographic texture. Each native metric is judged on what it claims to measure: motion smoothness on continuity, an action classifier on whether the named action occurs.

\section{CodingVBench}
\label{sec:audit}

\begin{figure*}[ht]
\centering
\includegraphics[width=\textwidth]{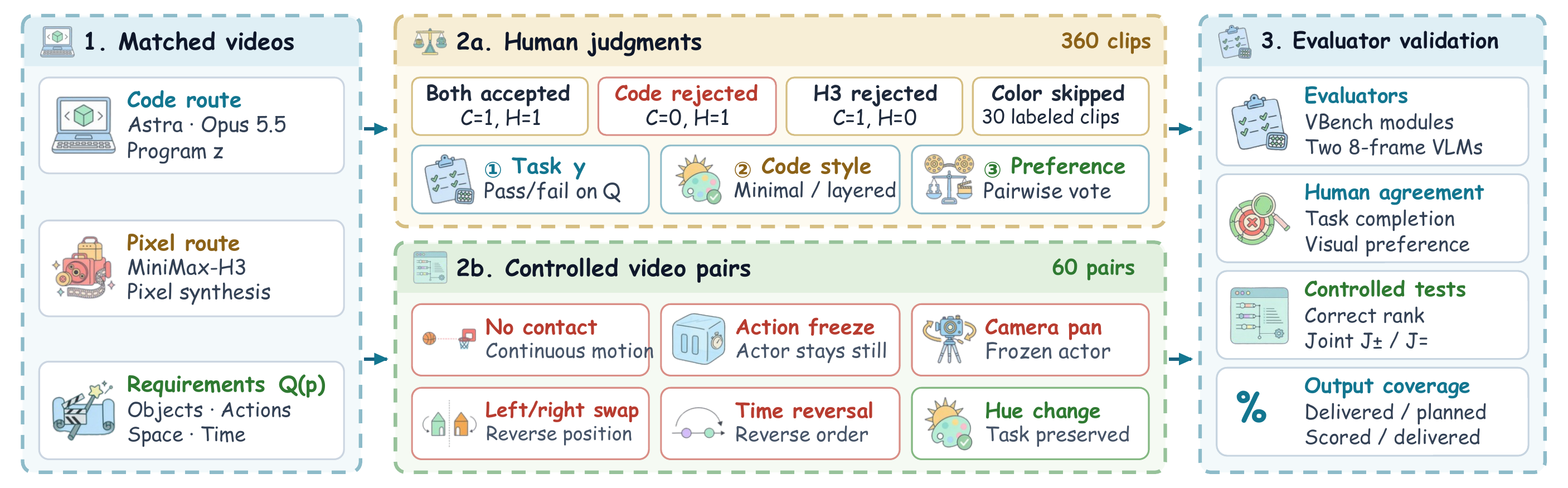}
\caption{\textbf{Fixed videos with human labels and verified edits test automatic evaluators.} Left: both routes answer $Q(p)$. Center top: three annotators assess 360 matched videos from Astra, Opus 5.5, and MiniMax-H3; $C/H$ select the original pairs and $y$ records human completion. A separate diagnostic set contains 30 labeled skipped-color clips. Center bottom: 50 behavioral-violation and ten task-preserving hue pairs. Right: native modules and two eight-frame VLM judges are compared with humans and on controlled pairs.}
\label{fig:method}
\end{figure*}

CodingVBench tests video evaluators on fixed videos with human labels and verified behavioral changes (Figure~\ref{fig:method}). Its data are 360 matched videos for 120 VBench prompts, 30 color clips that a native detector skipped, and 60 controlled video pairs. The matched videos carry completion, code-style, and pairwise visual-preference labels. The pairs contain 50 behavioral violations and ten task-preserving hue edits. The broader generation study covers 944 VBench prompts, including a 110-prompt 3D comparison (Section~\ref{sec:setup}). VBench supplies the prompts, and we add generated videos, annotations, and controlled variants. Appendix~\ref{app:reproducibility} describes the data release and the framework for evaluating new outputs.

\subsection{Human Judgments}
\label{sec:human}
Each prompt has one GPT-6 Astra, one Claude Opus 5.5, and one MiniMax-H3 video. Before annotation, we choose prompts by how the native score treats the original code--pixel pair: both videos accepted (60 prompts), only the pixel video accepted (40), and only the code video accepted (20). We then add the other coding model's output, so both code models answer the same requests. This choice uses native scores only and does not depend on the human labels.

Three annotators independently watch each full video. For completion they mark pass or fail on all visible requirements in $Q(p)$, and unverifiable cases stay unresolved. For visual preference they compare composition, lighting, texture, and rendering detail, and choose code better, MiniMax-H3 better, or comparable. Code style, from minimal geometry to layered illustration, is a separate descriptive label. We use majority labels. Annotators agree well on pixel completion (Fleiss' $\kappa$ of 0.72) and moderately on code style (0.55--0.66) and visual preference (0.54--0.59). Code completion has a $\kappa$ of 0.34--0.42 because almost every code video passes, while raw pairwise agreement on it is 92--96\%. Appendices~\ref{app:annotation} and~\ref{app:cvb_style} give annotation details and style statistics.

A separate diagnostic set contains 30 delivered color clips without a native score, split equally between code and MiniMax-H3. GRiT detects no target object, so the color module omits these clips. Labels record whether the requested colored object appears and, for code clips, its style (Appendix~\ref{app:annotation}). The set has no paired preference vote and stays outside the matched completion comparison.

\subsection{Multi-Frame VLM-as-Judge}
\label{sec:cvb_judge}
We use frozen GPT-6.1 Sol as the completion judge. It receives the prompt and eight frames sampled uniformly across the video, including the first and last, each with its timestamp and a maximum side of 448 pixels. It decides whether the requested objects and attributes are present, the named action occurs, and any specified spatial or temporal relation holds. For an action defined by contact, a recognizable actor and prop are not enough unless the contact is visible. The same criterion applies to code-rendered and MiniMax-H3 videos.

Claude Sonnet 5.5 receives the same instructions and frames as a second judge. The frame-count study keeps the primary judge and its decision rule fixed and varies only the number of frames (Section~\ref{sec:frames}).

\subsection{Controlled Video Comparisons}
\label{sec:interventions}
Human agreement checks the verdict on a video that a model happened to produce, but does not reveal which visual evidence drives that verdict. Controlled pairs address this limitation by changing one aspect of a video while keeping the remaining scene largely unchanged. Starting from a visually verified video $v^+$ that satisfies all requirements in $Q(p)$, we construct
\begin{equation}
(v^+,v^-),\qquad (v^+,v^=),
\label{eq:cvb_pairs}
\end{equation}
where $v^-$ violates one requirement and $v^=$ keeps every requirement after a whole-frame hue shift. Rejecting $v^-$ tests error detection, and accepting $v^=$ tests whether a correct decision survives an irrelevant hue shift.

The 60 pairs comprise 50 violating pairs, ten for each of five edits, and ten hue pairs. ``Non-contact motion'' redirects a moving object away from its required target. ``Action freeze'' holds the actor still. ``Camera pan'' moves the viewpoint across a frozen action, which separates camera movement from action execution. ``Left/right swap'' reverses the requested spatial relation, and ``Time reversal'' reverses the event or camera direction. ``Hue change'' shifts color across the frame and keeps every requirement of the prompt. We inspect each rendered pair to confirm the intended change and detect unintended occlusion or other visible effects. Paired videos share duration, resolution, and frame rate.

\subsection{Agreement, Paired Decisions, and Coverage}
\label{sec:metrics}
A native score becomes an accept or reject decision, and we call a decision correct when it matches the human label. Agreement is the share of correct decisions among the videos the evaluator scores, over both routes together or over one route. We also report the share of human-completed videos the evaluator accepts or rejects, and the share of human-failed videos it accepts. Appendix~\ref{app:cvb_error_rates} gives the exact definitions.

Delivery and scoring are counted separately. A video can be missing because the model never delivered it, or delivered but left unscored by a native detector. We therefore report delivered videos over planned outputs, and scored videos over delivered ones (Appendix~\ref{app:cvb_aggregation}).

On a controlled pair, an evaluator ranks correctly when it scores the completed original above the edited video. A tie is a failure to separate them. A hue pair has no preferred order because both videos complete the request. A correct ranking does not mean the decisions are right, since a judge can score the original higher and still reject both. For violation pairs we therefore also count the pairs where the judge accepts the original and rejects the edited video, and for hue pairs the pairs where it accepts both. We call this the both-decisions-right rate. Raw score changes are in the appendix.

\section{Experiments}
\label{sec:experiments}

We first compare completion decisions and aesthetic rankings with their human references, then test how the evaluators respond to controlled behavior changes and to the number of input frames, and finally look at how delivery and coverage affect model comparisons.

\subsection{Evaluation Setup}
\label{sec:setup}
\paragraph{Videos.}
The full evaluation uses 944 unique VBench prompts across all 16 dimensions~\cite{huang2024vbench}. Four coding models (Claude Opus 4.8~\cite{anthropic2026opus48}, Claude Opus 5.5~\cite{anthropic2026opus55}, GPT-6.1 Sol~\cite{openai2026gpt61sol}, and GPT-6 Astra~\cite{openai2026gpt6}) generate HTML scene programs, and MiniMax-H3 is the pixel-generation baseline. On a stratified 110-prompt subset we also compare direct HTML generation with multi-turn Three.js and Blender agents. Rendering settings and the restriction on text labels are in Appendix~\ref{app:rendering}. The human audit uses code videos from GPT-6 Astra and Claude Opus 5.5, and all controlled originals come from GPT-6 Astra.

\paragraph{Human audit and judges.}
The matched audit covers the action, spatial, object, and color dimensions. It yields 359 completion labels (239 code, 120 pixel), and one code video stays unresolved. Each evaluator is assessed on the labeled videos it scores. Each MiniMax-H3 video is counted once for completion, although it takes part in two code--pixel preference comparisons. Native UMT/GRiT decisions are compared with completion labels and LAION aesthetic scores with preferences.

\paragraph{Controlled edits and full-suite scores.}
The controlled set contains 50 behavioral comparisons, ten per intervention family, and ten hue comparisons. We report ranking correctness and the both-decisions-right rate on each evaluator's original score scale, distinguishing relative discrimination from correct binary decisions. Full-suite scores either assigning zero to undelivered videos or excluding them. Detector abstentions are tracked separately to distinguish generation failures from missing evaluation scores.

\subsection{Agreement with Human Completion}
\label{sec:alignment}
Both VLM judges agree with human completion far more often than the native semantic modules (Table~\ref{tab:alignment}). Overall agreement rises from 71.6\% to 92.7\% with GPT-6.1 Sol and 93.9\% with Claude Sonnet 5.5. The native modules make different errors on the two routes: they falsely reject 26.1\% of completed code videos, and they accept 75.0\% of failed MiniMax-H3 videos. Sol lowers these rates to 1.3\% and 37.5\%. Sonnet rejects none of the completed code videos but accepts 43.8\% of the failed pixel videos.

Most code videos are completed, so agreement alone separates evaluators less than it appears to. Of 239 code videos, 230 are completed, and a rule that accepts every video agrees with 96.2\% of the code labels and 88.6\% of all labels. We therefore read the judges through their two error types. Sol's gain on code videos comes from accepting completed animations: it accepts eight of the nine failed code videos, against six for the native modules. On the color dimension, Sol also scores 99.6\% of delivered code clips and 100\% of MiniMax-H3 clips, where the native detector scores 84.4\% and 76.9\%.

\begin{figure}[ht]
\centering
\vspace{1mm}
\includegraphics[width=\columnwidth]{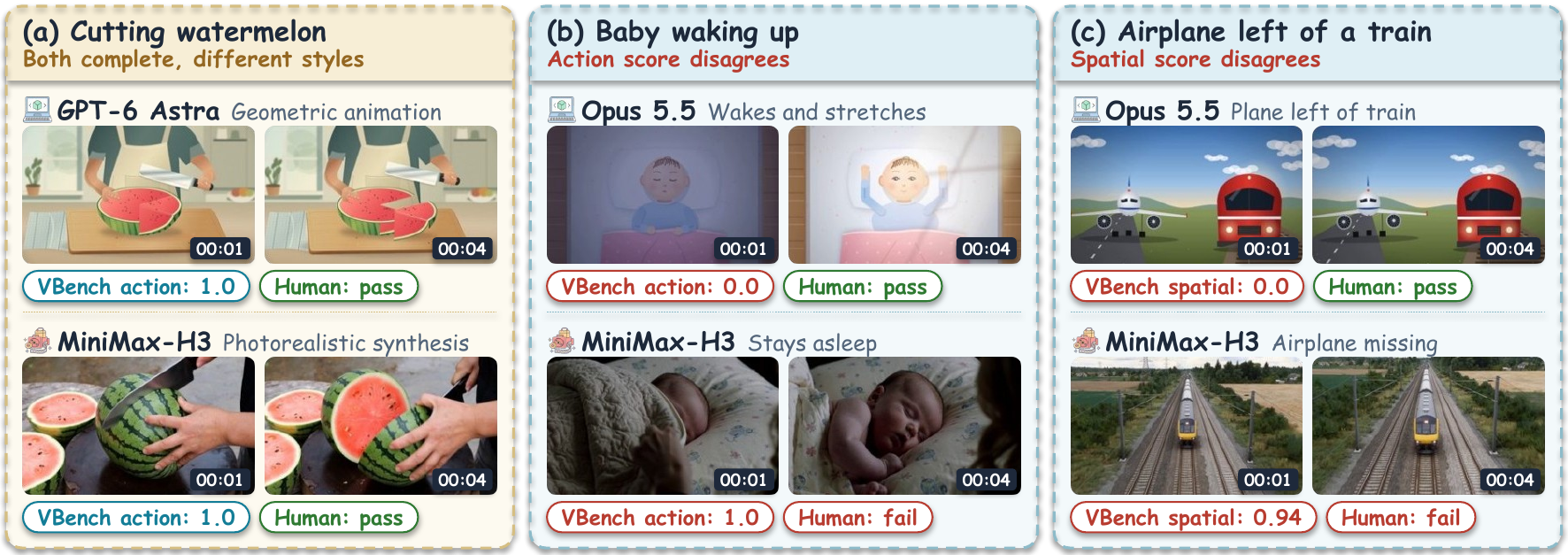}
\vspace{1mm}
\caption{\textbf{VBench can reject completed requests and accept failures.} (a) Both watermelon videos (code: GPT-6 Astra) complete the action and score 1.0 despite different styles. (b,c) Compliant code videos (Claude Opus 5.5) receive low scores, while MiniMax-H3 videos missing the event or object receive high scores. Frames are shown at one and four seconds. Action or spatial scores and human completion labels are listed separately.}
\label{fig:actions}
\end{figure}

Figure~\ref{fig:actions} shows the native errors directly. The code-rendered baby wakes and stretches but receives an action score of zero, and the MiniMax-H3 baby stays asleep and receives one. A code-rendered airplane correctly placed left of a train scores zero, while a MiniMax-H3 train without the airplane receives a high spatial score. In both cases the score ignores the event or relation that the prompt asks for.

The judges' decisions remain stable across coding models (Appendix~\ref{app:judges}). On the same 120 prompts, Sol passes 95.7\%--99.2\% of the videos of each of the four coding models, Sonnet passes 95.7\%--100\%, and the two judges agree with each other on 96.5\%--99.2\% of videos. Native VBench accepts 58.8\% and 62.6\% of the Opus 5.5 and Opus 4.8 videos but 86.8\% and 87.5\% of the Sol and Astra videos, a substantial gap that neither judge shows. This suggests that native acceptance is more sensitive to differences in rendering style across coding models. 

\begin{table*}[ht]
\centering
\footnotesize
\setlength{\tabcolsep}{3pt}
\caption{\textbf{The VLM judges agree with human completion more often than native modules on both routes.} (a) Agreement is the share of decisions that match the human completion label. The reference set has 359 videos (239 code, 120 MiniMax-H3), of which nine code and 32 pixel videos failed; each evaluator is measured on the videos it scores. The last four columns split the decisions by the human label: completed videos accepted or rejected, and failed videos accepted. The reference row accepts every video. (b) Coverage is scored over delivered color clips. All rates are percentages, and the best evaluator in each column is in bold.}
\label{tab:alignment}
\par\smallskip
\textbf{(a) Completion judgments}\par\smallskip
\begin{tabular*}{\textwidth}{@{\extracolsep{\fill}}lrrrrrrr@{}}
\toprule
Evaluator & \shortstack{Overall\\agreement $\uparrow$} & \shortstack{Code\\agreement $\uparrow$} & \shortstack{Pixel\\agreement $\uparrow$} & \shortstack{Code done,\\accepted $\uparrow$} & \shortstack{Code done,\\rejected $\downarrow$} & \shortstack{Pixel done,\\accepted $\uparrow$} & \shortstack{Pixel failed,\\accepted $\downarrow$} \\
\midrule
\textit{Accept every video (reference)} & \textit{88.6} & \textit{96.2} & \textit{73.3} & \textit{100.0} & \textit{0.0} & \textit{100.0} & \textit{100.0} \\
\midrule
VBench (UMT/GRiT) & 71.6 & 72.4 & 70.0 & 73.9 & 26.1 & 86.4 & 75.0 \\
VLM judge, GPT-6.1 Sol (8 frames) & 92.7 & 95.4 & 87.4 & 98.7 & 1.3 & 96.6 & \textbf{37.5} \\
VLM judge, Claude Sonnet 5.5 (8 frames) & \textbf{93.9} & \textbf{96.7} & \textbf{88.3} & \textbf{100.0} & \textbf{0.0} & \textbf{100.0} & 43.8 \\
\bottomrule
\end{tabular*}
\par\medskip\textbf{(b) Color scoring coverage}\par\smallskip
\begin{tabular*}{\textwidth}{@{\extracolsep{\fill}}lrr@{}}
\toprule
Evaluator & Code coverage $\uparrow$ & Pixel coverage $\uparrow$ \\
\midrule
VBench (GRiT) & 84.4 & 76.9 \\
GPT-6.1 Sol (8 frames) & \textbf{99.6} & \textbf{100} \\
\bottomrule
\end{tabular*}
\end{table*}

\subsection{Completion, Style, and Visual Preference}
\label{sec:cvb_visual_results}
Two coding models with nearly the same human completion rate receive very different native acceptance (Table~\ref{tab:stratified}). On the same 120 prompts, humans judge 95.8\% of GPT-6 Astra videos and 96.6\% of Claude Opus 5.5 videos complete (115 of 120 and 115 of 119 resolved), while native VBench accepts 87.5\% and 59.2\%. It falsely rejects 12.2\% of Astra's completed videos and 40.0\% of Opus 5.5's.

The model with the lower native acceptance also draws more simply. Minimal geometric rendering accounts for 46.7\% of Opus 5.5 videos and 29.2\% of Astra videos, and humans prefer MiniMax-H3's visual quality on 51.7\% of Opus 5.5 pairs against 31.7\% of Astra pairs. Native acceptance thus follows how elaborate the rendering is, while human completion stays about the same. Because independently generated clips also differ in motion, framing, and object appearance, the next section isolates single factors with controlled edits. The same pattern appears across the three groups of prompts we chose by native score (Appendix~\ref{app:cvb_style}): minimal geometry is 82.5\% of the code videos that native modules reject, against 33.3\% and 25.0\% in the other two groups, and humans still judge 92.3\% of those rejected videos complete.

Native aesthetic rankings often oppose human preference. The predictor scores the code video higher in 196 of the 239 pairs with both scores. Among the 99 scored pairs where humans prefer MiniMax-H3, it still favors code in 81 (82\%). On the same videos, the score favors the output that humans judged worse.

\begin{table*}[ht]
\centering
\small
\setlength{\tabcolsep}{4pt}
\caption{\textbf{Similar completion, different native acceptance and style.} Both models receive the same 120 prompts. Human completion is the share of resolved videos labeled complete, and VBench accepts is the share accepted among natively scored videos. False rejection uses human-completed videos as its denominator. Minimal style is the share of labeled code videos that use minimal geometry. H3 preference uses all labeled pairs, including comparable outcomes. Rates are percentages, and the aesthetic columns average the available VBench LAION scores.}
\label{tab:stratified}
\begin{tabular*}{\textwidth}{@{\extracolsep{\fill}}lrrrrrrr@{}}
\toprule
Code model & \shortstack{Human\\completion} & \shortstack{VBench\\accepts} & \shortstack{Native false\\rejection} & \shortstack{Minimal\\style} & \shortstack{H3 visual\\preference} & \shortstack{Code\\aesthetic} & \shortstack{H3\\aesthetic} \\
\midrule
GPT-6 Astra & 95.8 & \textbf{87.5} & \textbf{12.2} & \textbf{29.2} & \textbf{31.7} & \textbf{63.9} & \textbf{53.8} \\
Claude Opus 5.5 & \textbf{96.6} & 59.2 & 40.0 & 46.7 & 51.7 & 60.9 & 53.7 \\
\bottomrule
\end{tabular*}
\end{table*}

\subsection{Responses to Controlled Behavior Changes}
\label{sec:controlled}
Native action scores often stay the same when the requested action is removed (Table~\ref{tab:controlled}). All ten variants with continuous motion but no defining contact tie with their completed originals, and freezing the actor or panning the camera across a frozen pose ties in 90\% of pairs. The native action score accepts a recognizable scene even when the interaction that defines the action is absent.

\begin{figure}[ht]
\centering
\includegraphics[width=\columnwidth]{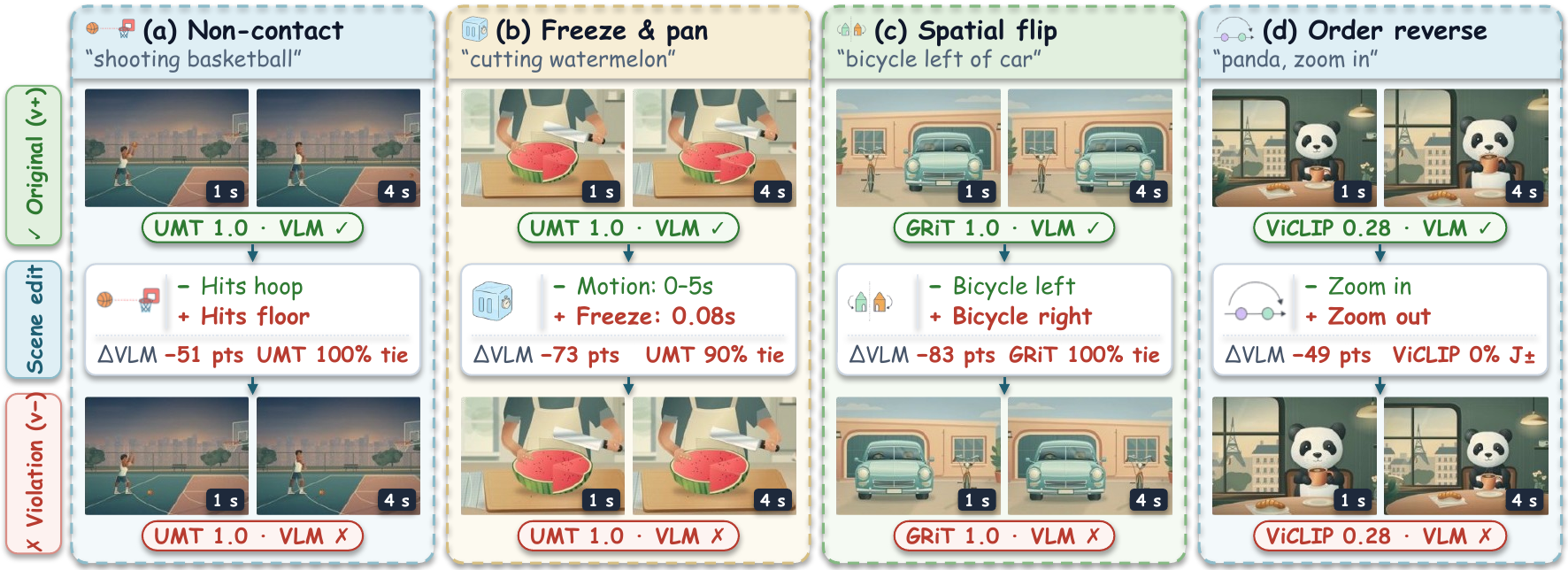}
\caption{\textbf{Removing required behavior can leave native scores unchanged.} Frames at one and four seconds compare the verified original $v^+$ with a violation $v^-$. Scene edits change (a) basketball contact, (b) action execution, (c) left-right placement, or (d) zoom direction. Native scores change little or tie. $\Delta$VLM is GPT-6.1 Sol's violation score minus its original score, on its original scale. Green and coral mark completed and violating videos. }
\vspace{-2mm}
\label{fig:timing}
\end{figure}

The spatial and temporal edits show the same pattern (Figure~\ref{fig:timing}). All ten left-right reversals keep the original spatial score. The rule we inspected compares unsigned distances and assigns detections to objects inconsistently (Appendix~\ref{app:spatial-rule}), and horizontal reflection leaves those distances unchanged. Reversing a zoom moves the ViCLIP score only from 0.2854 to 0.2861, although the requested camera direction is reversed.

GPT-6.1 Sol ranks all 50 behavioral violations below their completed originals. Counting only the pairs where the judge accepts the original and rejects the edited video is stricter, and gives between 70\%-100\% across the five families. For non-contact motion, all ten originals outrank their violations, but only seven pairs get both decisions right. A higher score for the original gives the right order, and the judge must also put the accept/reject line between the two videos. Sonnet 5.5 ranks 46/50 pairs correctly, and gets both decisions right in 40\% of non-contact pairs and 60\% of camera pan pairs. Sonnet leads in agreement with humans and Sol leads on the controlled pairs, so the two tests rank the judges differently. Appendix~\ref{app:judges} gives detailed results.

The hue pairs ask whether a completed video stays accepted after an irrelevant color change. Sol accepts both videos in nine of ten pairs. In the tenth, it rejects the original as well as the edited video (Appendix~\ref{app:vlm_errors}), so the hue shift did not cause the error. Native UMT accepts both videos in every hue pair, and it also accepts every non-contact violation. A judge that accepts everything is perfectly stable under color and blind to behavioral failure, which is why the two tests are read together.

\begin{table}[ht]
\centering
\footnotesize
\setlength{\tabcolsep}{3pt}
\caption{\textbf{The VLM detects behavioral changes that native scores miss.} Each family contains ten pairs. Cells report VBench / GPT-6.1 Sol, and the Both right column adds Claude Sonnet 5.5 as a third value. Correct rank means the completed original scores above its violation. Both right means the judge accepts the original and rejects the edited video (for hue, accepts both). Equal scores are descriptive, and hue pairs have no preferred ordering. All entries are percentages; raw score changes are in the appendix.}
\label{tab:controlled}
\begin{tabular*}{\columnwidth}{@{\extracolsep{\fill}}lrrr@{}}
\toprule
Intervention & Correct rank $\uparrow$ & Equal scores & Both right $\uparrow$ \\
\midrule
Action freeze & 10 / \textbf{100} & 90 / \textbf{0} & 10 / \textbf{90} / \textbf{90} \\
Non-contact motion & 0 / \textbf{100} & 100 / \textbf{0} & 0 / \textbf{70} / 40 \\
Camera pan & 10 / \textbf{100} & 90 / \textbf{0} & 10 / \textbf{90} / 60 \\
Spatial flip & 0 / \textbf{100} & 100 / \textbf{0} & 0 / \textbf{100} / \textbf{100} \\
Temporal reverse & 40 / \textbf{100} & 0 / 0 & 0 / \textbf{100} / 90 \\
Hue change & -- & 100 / 40 & \textbf{100} / 90 / \textbf{100} \\
\bottomrule
\end{tabular*}
\end{table}

\subsection{Effect of Frame Count}
\label{sec:frames}
Eight frames matter most when completion depends on change over time (Table~\ref{tab:frames}). We vary the input to GPT-6.1 Sol and keep its instructions and decision rule fixed. Agreement with human completion is 91.1\% with one middle frame, 92.5\% with four frames, and 92.7\% with eight, so it barely moves. The controlled pairs show a much larger difference: with one frame the judge gets both decisions right in only 20\% of action freeze pairs, 20\% of camera pan pairs, and 0\% of temporal reversal pairs, and eight frames raise these to 90\%, 90\%, and 100\%. A still image cannot show whether an actor moves or whether a camera sequence runs in the requested direction. In contrast, spatial reversal reaches 100\% at every frame count because left and right relationships are directly visible in one frame.

Recognizing objects in a still image does not show that the requested transition occurred, and agreement over the whole audit hides this. Temporal pairs are therefore needed to test a completion judge. Four frames recover temporal reversal, but non-contact pairs stay at 40\% and reach 70\% only with eight, so the number of frames should follow the event being checked: spatial layout needs one, and contact and temporal change need a sequence.

\begin{table*}[ht]
\centering
\footnotesize
\setlength{\tabcolsep}{2.5pt}
\caption{\textbf{Temporal tests benefit most from multiple frames.} GPT-6.1 Sol uses one middle frame or four or eight uniformly spaced frames with the same decision rule. Agreement is measured against human completion labels on the 359 labeled videos; Overall covers both routes and Pixel uses pixel videos only. ``Code done, rejected'' is the share of human-completed code videos the judge rejects, and ``Pixel failed, accepted'' the share of human-failed pixel videos it accepts. Right: the both-decisions-right rate over ten pairs per family, which requires accepting the original and rejecting the edited video, or accepting both videos for hue. All rates are percentages, and the best are in bold.}
\label{tab:frames}
\begin{tabular*}{\textwidth}{@{\extracolsep{\fill}}lrrrrrrrrrr@{}}
\toprule
 & \multicolumn{4}{c}{Against human completion labels} & \multicolumn{6}{c}{Controlled pairs, both decisions right $\uparrow$} \\
\cmidrule(lr){2-5}\cmidrule(lr){6-11}
Frames & \shortstack{Overall\\agreement $\uparrow$} & \shortstack{Code done,\\rejected $\downarrow$} & \shortstack{Pixel failed,\\accepted $\downarrow$} & \shortstack{Pixel\\agreement $\uparrow$} & \shortstack{Action\\freeze} & \shortstack{Non-contact\\motion} & \shortstack{Camera\\pan} & \shortstack{Spatial\\flip} & \shortstack{Temporal\\reverse} & \shortstack{Hue\\change} \\
\midrule
1 & 91.1 & 1.3 & 43.8 & 82.5 & 20 & 40 & 20 & \textbf{100} & 0 & 60 \\
4 & 92.5 & 1.3 & \textbf{37.5} & 86.7 & 70 & 40 & 70 & \textbf{100} & \textbf{100} & 60 \\
8 & \textbf{92.7} & 1.3 & \textbf{37.5} & \textbf{87.4} & \textbf{90} & \textbf{70} & \textbf{90} & \textbf{100} & \textbf{100} & \textbf{90} \\
\bottomrule
\end{tabular*}
\end{table*}

\subsection{Remaining Completion Errors}
\label{sec:remaining_errors}
\begin{figure}[ht]
\centering
\includegraphics[width=\columnwidth]{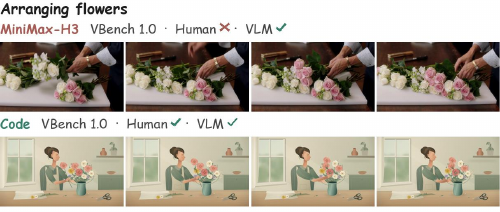}
\caption{\textbf{VLM acceptance can disagree with human completion.} For ``arranging flowers,'' MiniMax-H3 (top) fails the human criterion but is accepted by GPT-6.1 Sol, and the code video (bottom) passes both. Both receive VBench action scores of 1.0. Frames at 8\%, 35\%, 65\%, and 95\% of each clip illustrate the videos, not the eight-frame judge input.}
\label{fig:vlm_errors}
\end{figure}

The eight-frame Sol judge still accepts 12 of 32 failed MiniMax-H3 videos, with four frames yielding the same 37.5\% false acceptance rate. Figure~\ref{fig:vlm_errors} shows ``arranging flowers'': VBench and Sol accept both videos, while humans accept only the code-rendered one. Appendix~\ref{app:vlm_errors} discusses sampled-frame limitations in detecting such errors.

\subsection{Delivery and Scoring Coverage}
\label{sec:baseline}
Which videos enter an average changes the ranking of the systems (Table~\ref{tab:baseline}). With undelivered videos scored as zero, MiniMax-H3 ranks above the coding models, and with them excluded, Astra and Sol rank above MiniMax-H3 (Appendix~\ref{app:cvb_aggregation}). A reported score should say which planned outputs it averages.

Native detectors also leave delivered videos unscored. Of Opus 5.5's 425 planned color videos, 423 are delivered and 357 are scored. All 15 skipped code videos contain the requested colored object, and six of the 15 skipped MiniMax-H3 videos do not, so dropping abstentions removes both completed requests and failures.

\begin{table}[ht]
\centering
\footnotesize
\setlength{\tabcolsep}{3pt}
\caption{\textbf{Counting undelivered videos changes model comparisons.} Full-suite totals cover 944 prompts and assign zero to undelivered outputs. Delivery is a percentage over 6,220 planned outputs per system, and scores use the 0--100 VBench scale. The best coding-model score is in bold. Totals that exclude undelivered videos, and the missing counts, are in Appendix~\ref{app:cvb_aggregation}.}
\label{tab:baseline}
\begin{tabular*}{\columnwidth}{@{\extracolsep{\fill}}lrr@{}}
\toprule
System & Missing$=0$ $\uparrow$ & Delivered $\uparrow$ \\
\midrule
MiniMax-H3 & 80.51 & 100.0 \\
Claude Opus 5.5 & \textbf{78.62} & 98.0 \\
GPT-6 Astra & 77.25 & 93.7 \\
GPT-6.1 Sol & 77.21 & 94.6 \\
Claude Opus 4.8 & 75.66 & 97.9 \\
\bottomrule
\end{tabular*}
\end{table}

\section{Discussion and Limitations}
\label{sec:discussion}

Agreement with humans and the response to controlled edits answer different questions, and both should inform the choice of a completion judge. Human labels check decisions on real model outputs, and controlled pairs expose missed behaviors that agreement alone hides. Multiple frames supply the change over time that separates an executed action from a still pose.

The hue pairs test stability under one color change. Shading, texture, and geometry need their own task-preserving comparisons. Human labels cover two coding models and one pixel generator, the other coding models' results are automatic judgments (Appendix~\ref{app:judges}), and the controlled originals come from GPT-6 Astra. Longer narratives, physical interactions, and new generators need new labeled outputs and verified edits. The fixed videos and labels serve as test cases for comparing frozen judges and for developing future completion evaluators.

\section{Conclusion}
\label{sec:conclusion}

CodingVBench tests video evaluators using human-labeled outputs from two production routes and controlled pairs that isolate individual changes. Our results reveal weaknesses in native video metrics: they can rank failed videos above completed ones, overlook required behaviors, and favor videos that humans rate lower in visual quality. Multi-frame VLM judges achieve substantially higher agreement with human completion, while controlled interventions reveal whether their decisions respond to task-relevant behavior rather than incidental appearance. Reliable cross-route evaluation should therefore assess task completion, visual quality, delivery, and scoring coverage separately. CodingVBench provides a reproducible foundation for diagnosing evaluators and developing more reliable video evaluation methods across generation paradigms.


\label{maintext:end}
\typeout{CVPR_MAIN_END_PAGE=\thepage}

\clearpage
\section{LLM Usage Statement}

Large language models assisted with manuscript polishing, translation, and figure layout scripts. All experimental measurements derive from verified project runs. The authors take full responsibility for all data, code, and scientific claims.

{\small
\bibliographystyle{ieeenat_fullname}
\bibliography{main}
}
\appendix
\onecolumn
\pagenumbering{arabic}
\setcounter{page}{1}
\makeatletter
\@ifundefined{c@cvprrulercount}{}{\setcounter{cvprrulercount}{1}}
\makeatother

\begin{center}
    \Large \textbf{\thetitle}\\
    \vspace{0.5em}Supplementary Material \\
    \vspace{1.0em}
\end{center}

\section{Aggregation and Coverage}
\label{app:cvb_aggregation}
Table~\ref{tab:cvb_aggregation_app} averages the same outputs in two ways. ``Undelivered zero'' gives a missing video a score of zero, and ``Valid video'' leaves it out. The ranking of the systems differs between the two. Whether a judge's decisions are right is a separate question, answered in Section~\ref{sec:alignment}.

\begin{table}[ht]
\centering
\small
\caption{\textbf{Excluding undelivered videos changes the ranking.} Both columns use the same full-suite outputs. Drop is Valid video minus Undelivered zero, and Missing counts undelivered videos.}
\label{tab:cvb_aggregation_app}
\begin{tabular*}{\columnwidth}{@{\extracolsep{\fill}}lrrrr@{}}
\toprule
System & Valid video $\uparrow$ & Undelivered zero $\uparrow$ & Drop & Missing \\
\midrule
MiniMax-H3 & 80.51 & 80.51 & 0.00 & 0 \\
Claude Opus 4.8 & 78.00 & 75.66 & 2.34 & 131 \\
Claude Opus 5.5 & 80.82 & 78.62 & 2.20 & 127 \\
GPT-6.1 Sol & 83.23 & 77.21 & 6.02 & 334 \\
GPT-6 Astra & 83.36 & 77.25 & 6.11 & 391 \\
\bottomrule
\end{tabular*}
\end{table}

\subsection{Static Candidate Filtering}
For the temporal flickering dimension, VBench selects up to five static clips from 25 candidates and averages only those. We also look at the first five candidates to see how often a system follows a still-frame request such as ``In a still frame, a pear'' (Table~\ref{tab:cvb_filter_app}). Only 48\% (20/42) of the primary Three.js outputs and 51\% (22/43) of the primary Blender outputs from Claude Opus 4.8 agents are static, against 94\%--100\% for direct HTML and Codex agents. Selecting from 25 candidates fills nearly all static slots and gives flickering scores above 99.50. Restricting the evaluation to the first five candidates lowers the Three.js agent's total from 81.49 to 71.97, so the selected set hides how often the system disobeys the instruction.

\begin{table}[ht]
\centering
\small
\caption{\textbf{Static filtering hides moving outputs.} Primary static is static over delivered among the first five candidates per prompt. Kept counts static clips retained after selection from 25 candidates. The score columns are full totals for the 110-prompt subset under each candidate limit.}
\label{tab:cvb_filter_app}
\begin{tabular*}{\columnwidth}{@{\extracolsep{\fill}}lrrrr@{}}
\toprule
Configuration & Primary static & Kept & 25-candidate score & 5-candidate score \\
\midrule
HTML, Opus 5.5 & 47/50 & 50 & 81.18 & 78.52 \\
HTML, Astra & 44/44 & 50 & 81.44 & 77.45 \\
Three.js, Opus 4.8 & 20/42 & 46 & 81.49 & 71.97 \\
Blender, Opus 4.8 & 22/43 & 49 & 80.13 & 72.11 \\
Three.js, Astra & 48/50 & 50 & 84.00 & 84.00 \\
\bottomrule
\end{tabular*}
\end{table}

\clearpage
\subsection{Detector Coverage}
Among Opus 5.5 color videos, 357 of 423 delivered clips receive a native score, which is 84.4\% of delivered and 84.0\% of the 425 planned outputs (Table~\ref{tab:cvb_color_app}). Labels confirm the correct colored object in all 15 skipped code clips, and six of the 15 skipped MiniMax-H3 clips omit it. Thirteen of the 15 skipped code clips use non-minimal rendering, so the omissions are not limited to minimal geometric style.

\begin{table}[ht]
\centering
\small
\caption{\textbf{Delivery and scoring coverage have different denominators.} Scored/delivered is the share of delivered clips that receive a native score, and scored/planned also counts missing outputs. Coverage ratios are percentages. Conditional mean averages native color scores over scored clips, and Abstention zero averages over all delivered clips with unscored clips counted as zero.}
\label{tab:cvb_color_app}
\begin{tabular*}{\columnwidth}{@{\extracolsep{\fill}}lrrrrrrr@{}}
\toprule
System & Planned & Delivered & Scored & \shortstack{Scored/\\delivered} & \shortstack{Scored/\\planned} & \shortstack{Conditional\\mean} & \shortstack{Abstention\\zero} \\
\midrule
Opus 5.5 & 425 & 423 & 357 & 84.4 & 84.0 & 89.31 & 75.38 \\
MiniMax-H3 & 425 & 425 & 327 & 76.9 & 76.9 & 75.49 & 58.08 \\
\bottomrule
\end{tabular*}
\end{table}

\subsection{Agreement and Completion Error Rates}
\label{app:cvb_error_rates}
For a video that an evaluator scores, a decision is correct when it matches the human completion label. The rates in the main text are
\begin{itemize}[leftmargin=*,itemsep=0pt]
\item \emph{agreement}: correct decisions over all scored videos with a human label;
\item \emph{completed, accepted}: accepted videos over scored videos that humans labeled completed;
\item \emph{completed, rejected}: rejected videos over the same videos, so it equals 100\% minus the previous rate;
\item \emph{failed, accepted}: accepted videos over scored videos that humans labeled failed.
\end{itemize}
Overall agreement counts the scored code and pixel videos together and does not average the two route percentages. The route-specific rates use only that route's videos. Human completion is the share of labeled videos that are completed, and a judge's pass rate is the share of scored videos it accepts; neither is agreement. Color coverage divides by delivered clips and is separate from correctness.

\section{Human Annotation Agreement}
\label{app:annotation}
Three annotators label the 360 matched videos, and one code completion stays unresolved after aggregation, which leaves 359 labels. The separate 30-clip skipped-color set has three-annotator majority labels for 29 videos (14 code, 15 MiniMax-H3). The remaining code video uses an author-provided label that the requested colored object is present and the rendering is non-minimal. These labels are not part of the matched completion comparison. Visual preference has three outcomes: code better, MiniMax-H3 better, and comparable.

On the matched videos, Fleiss' $\kappa$ is 0.34--0.42 for code completion across two annotation rounds, 0.72 for pixel completion, 0.55--0.66 for code style, and 0.54--0.59 for visual preference. Raw pairwise agreement on code completion is 92--96\%. Most code videos satisfy the request, which keeps chance-corrected agreement lower than raw agreement.

The final completion reference has 230 completed and nine failed code videos, and 88 completed and 32 failed MiniMax-H3 videos. Accepting every video would therefore agree with 318 of 359 labels (88.6\%), 230 of 239 code labels (96.2\%), and 88 of 120 pixel labels (73.3\%). This is the reference row of Table~\ref{tab:alignment}. Because so few code videos fail, an evaluator can raise agreement by rejecting less without detecting more failures. We report acceptance of completed and failed videos separately, and use the controlled pairs to test whether a judge notices a removed behavior.

\clearpage
\section{Judge Comparison Across Coding Models}
\label{app:judges}
Table~\ref{tab:judges_app} applies both VLM judges to the code videos of all four coding models on the 120 prompts. Human labels exist only for Astra and Opus 5.5, so the other rows show automatic decisions. The two judges agree with each other on 96.5\%--99.2\% of videos, and both pass 95.7\%--100\% of the videos of every model, while native VBench accepts far fewer videos from the Claude models. Counts differ slightly across rows because a few videos are missing or were not scored by a judge.

\begin{table}[ht]
\centering
\small
\caption{\textbf{Both VLM judges give similar decisions for every coding model.} Pass rates are percentages of judged videos, and VBench accepts means a native score of at least 0.5. Judges agree compares the two VLM decisions on jointly judged videos, independently of human labels. Opus 4.8 and Sol videos have no human labels.}
\label{tab:judges_app}
\begin{tabular*}{\columnwidth}{@{\extracolsep{\fill}}lrrrrr@{}}
\toprule
Coding model & Videos & \shortstack{VBench\\accepts} & \shortstack{GPT-6.1\\Sol passes} & \shortstack{Sonnet 5.5\\passes} & \shortstack{Judges\\agree} \\
\midrule
Claude Opus 4.8 & 117 & 62.6 & 95.7 & 95.7 & 98.3 \\
Claude Opus 5.5 & 119 & 58.8 & 99.2 & 100.0 & 99.2 \\
GPT-6.1 Sol & 114 & 86.8 & 96.5 & 98.2 & 96.5 \\
GPT-6 Astra & 120 & 87.5 & 97.5 & 99.2 & 98.3 \\
\bottomrule
\end{tabular*}
\end{table}

On the controlled pairs, the share of pairs where the judge accepts the original and rejects the edited video (accepts both, for hue) is, for GPT-6.1 Sol and Claude Sonnet 5.5 respectively, 90\% and 90\% for action freezing, 70\% and 40\% for non-contact motion, 90\% and 60\% for camera pan, 100\% and 100\% for left-right reversal, 100\% and 90\% for temporal reversal, and 90\% and 100\% for hue changes.

\section{Style and Visual Richness}
\label{app:cvb_style}
Table~\ref{tab:cvb_style_app} gives style and richness ratings for the three native-score groups used to choose prompts. Code videos that the native modules reject are more often minimal in style, although human completion stays high. In these groups the code video came from GPT-6 Astra for 94 prompts and from Claude Opus 5.5 for 26, so group and coding model overlap; Table~\ref{tab:stratified} compares the two models on identical prompts. The controlled edits, not these groups, test the effect of specific behavioral and color changes.

\begin{table}[ht]
\centering
\small
\caption{\textbf{Simpler code renderings can still complete the task.} Minimal style corresponds to the ``crude'' tier of the annotation. The skipped-color row covers 15 code and 15 pixel clips without native scores, not matched preference pairs, and its completion columns report presence of the requested colored object (Appendix~\ref{app:annotation}). Completion and style are percentages; richness and aesthetics use the available original scores.}
\label{tab:cvb_style_app}
\begin{tabular*}{\columnwidth}{@{\extracolsep{\fill}}lrrrrrr@{}}
\toprule
Group & \shortstack{Code\\completion} & \shortstack{H3\\completion} & \shortstack{Code\\minimal style} & \shortstack{Code\\richness} & \shortstack{H3\\richness} & \shortstack{Aesthetic\\Code / H3} \\
\midrule
Both accepted & 95.0 & 80.0 & 33.3 & 73.2 & 80.8 & 63.4 / 53.6 \\
Code rejected & 92.3 & 70.0 & 82.5 & 57.5 & 82.2 & 62.9 / 55.3 \\
H3 rejected & 95.0 & 60.0 & 25.0 & 74.5 & 80.8 & 62.7 / 51.0 \\
Color skipped & 100.0 & 60.0 & 13.3 & 52.1 & 81.1 & 61.4 / 57.6 \\
\bottomrule
\end{tabular*}
\end{table}

\clearpage
\section{Detailed Controlled Results}
\label{app:cvb_controls}
For a violation pair, the score change is the score of the edited video minus the score of the original; for a hue pair, the hue-shifted video replaces the edited one. Table~\ref{tab:cvb_controls_app} gives these changes together with ties, ranking, and the both-decisions-right rate. The spatial scoring rule is described in Appendix~\ref{app:spatial-rule}. With ten pairs per family, a 10/10 result has a 95\% Wilson lower bound of 72\%.

\begin{table}[ht]
\centering
\small
\caption{\textbf{Raw score changes accompany the paired decisions.} Each cell reports native VBench / GPT-6.1 Sol. Rank, tie, and both-right rates are percentages, and score changes stay on each evaluator's original scale.}
\label{tab:cvb_controls_app}
\begin{tabular*}{\columnwidth}{@{\extracolsep{\fill}}lrrrr@{}}
\toprule
Intervention & Correct rank & Tie rate & Both right & Score change \\
\midrule
Action freeze & 10 / 100 & 90 / 0 & 10 / 90 & $-0.10/-72.7$ \\
Non-contact motion & 0 / 100 & 100 / 0 & 0 / 70 & $0.00/-50.5$ \\
Camera pan & 10 / 100 & 90 / 0 & 10 / 90 & $-0.10/-67.7$ \\
Spatial flip & 0 / 100 & 100 / 0 & 0 / 100 & $0.00/-83.0$ \\
Temporal reverse & 40 / 100 & 0 / 0 & 0 / 100 & $+0.00/-48.7$ \\
Hue change & -- & 100 / 40 & 100 / 90 & $0.00/+1.0$ \\
\bottomrule
\end{tabular*}
\end{table}

\section{Supplementary Constraint Cases}
\label{app:cvb_constraints}
We also built 14 explicit text and timing tasks, each with a positive video and a negative one that breaks a single property. A pixel-level checker separates all pairs, including text substitutions, timing-window violations, small glyphs, and brief flashes. In a reference-conditioned price-editing example, MiniMax-H3 keeps ``399'' at 10.75 seconds when asked to change it to ``299'', while a scene-program edit changes the string and leaves the untouched frames unchanged. These examples concern explicit constraints and say nothing about general editing ability.

\section{Evaluator Adaptation and Case Studies}
\label{app:training}

\subsection{Potential Evaluator Adaptation}
The verified completion labels and controlled pairs could supervise a video-language completion evaluator that both accepts completed videos and rejects behavioral violations while staying stable under task-preserving edits. We do not train an evaluator in this paper.

\subsection{Curated Cross-Route Case Studies}
Inspection identified 29 diagnostic pairs with four recurring patterns. First, both routes can receive a perfect score despite a large realism gap: on ``cutting watermelon'' (replicate 0), GPT-6 Astra and MiniMax-H3 both score 1.0 and complete the action, although Astra uses geometric primitives. Second, a pixel-generated clip can receive a high score without the requested action or relation, as in ``clay pottery making'' and ``arranging flowers'' (replicate 0), or ``a couch on the right of a bed'' (replicate 1). Third, compliant code animations can score 0.0 while incomplete MiniMax-H3 clips receive 0.94--1.00, as in ``baby waking up,'' ``planting trees,'' and ``an airplane on the left of a train'' (replicate 4), and ``air drumming'' (replicate 2). Fourth, the metric separates some pairs correctly: code clips score 1.0 and pixel-generated clips score 0.0 on ``bending metal'' and ``bending back'' (replicate 0).

\subsection{Remaining VLM Errors}
\label{app:vlm_errors}
The eight-frame GPT-6.1 Sol judge accepts 12 of the 32 MiniMax-H3 videos that humans labeled incomplete. Figure~\ref{fig:vlm_errors} shows one case, ``arranging flowers,'' with the completed code video for the same prompt: native VBench and Sol accept both. ``Doing laundry'' follows the same pattern. The displayed frames show recognizable tools and activity, while the human labels come from the full videos, so a recognizable activity is not enough to show that the request is completed.

The hue pair for ``sword fighting'' shows a different error. The verified original and its hue-shifted version receive VLM scores of 48 and 52, and both are rejected because the judge reports no visible blade contact. Contact can fall between the sampled frames, and we have not separated sampling from the judge's interpretation. The original is rejected as well, so this is a completion error that the hue change did not cause.

\section{Reproducibility and Data Sources}
\label{app:reproducibility}

\paragraph{Benchmark data.}
The planned release contains the matched videos and their VBench prompts, per-video completion labels, code-style labels, and code--pixel preference votes. It also contains the 30 color videos that the native detector skipped, with their object-presence and code-style annotations, and the 60 controlled pairs with their original and edited videos, verified task status, scene programs, and edit descriptions. Completion labels exist for 359 of the 360 matched videos, and the unresolved case and the annotation sources for the color set are marked in the data (Appendix~\ref{app:annotation}). The controlled set is counted in pairs because originals can be reused, so 60 is not a count of unique videos.

\paragraph{Evaluation framework.}
We will provide the frame-sampling and VLM-judging instructions, the rules that turn scores into decisions, the evaluation code, and the per-video outputs behind our comparisons. An evaluator is tested by scoring the fixed videos and comparing its decisions with the human labels or with the known relation inside each pair. A new video model can instead generate videos for the shared prompts and be assessed on completion, visual quality, delivery, and coverage. The existing human labels apply only to the recorded videos.

\paragraph{Supporting generation study.}
The 944-prompt full-suite study and the 110-prompt 3D comparison give the broader generation context, with model versions, scores, and production records. The 40,100 in Figure~\ref{fig:overview} counts planned output slots, not delivered videos or the size of the labeled benchmark. Generation failures, rendering failures, detector abstentions, and unresolved judgments are recorded separately, so each result keeps its own denominator.

\subsection{VLM Input and Frame-Count Comparison}
\label{app:judge_frames}
GPT-6.1 Sol is the primary completion judge and the model used in the frame-count study. Claude Sonnet 5.5 receives the same eight-frame inputs and instructions as the second judge. For a video with $T$ decoded frames, an $n$-frame input uses the frames at indices
\begin{equation}
\operatorname{round}\!\left(\frac{i(T-1)}{n-1}\right),\quad i=0,\ldots,n-1,
\end{equation}
with $n=8$ by default, and the one-frame variant uses the middle frame. Each frame is resized to a longer side of at most 448 pixels, encoded as JPEG, and given its timestamp. The model and completion instructions are the same for one, four, and eight frames, and only the sentence stating the number of frames changes. The instructions ask whether the requested behavioral and semantic requirements are visibly realized and do not penalize non-photorealistic rendering unless the prompt asks for realism. They list incomplete or non-contacting actions, frozen actors, reversed spatial relations, and reversed camera direction as cases to check. The frame-count comparison uses the same 359 labeled videos and 60 controlled pairs as the main evaluation.

\subsection{Rendering Settings and Failure Analysis}
\label{app:rendering}

Code videos are rendered in headless Chromium with SwiftShader, and the same timestamp always produces the same frame. The playback interface takes a time $t$ and renders the corresponding scene state. Videos are $1280\times720$ at 24 fps and 5 seconds (120 frames), and on-screen labels that reveal the requested content are not allowed.

Of the 983 full-suite HTML failures, 493 (50.2\%) are JavaScript syntax errors or collisions with browser-global names, 172 (17.5\%) are CPU rendering timeouts, and 150 (15.3\%) are malformed color values in gradients. Per-frame blur filters appear in 93\% of the timed-out scenes and 32\% of controls. Figures~\ref{fig:exp-execution} and~\ref{fig:exp-timing} show the failure and runtime profiles. These failures decide whether a video is delivered and say nothing about the quality of the videos that rendered.

\subsection{Directional Spatial Scoring}
\label{app:spatial-rule}

The spatial rule we inspected compares the horizontal and vertical distances between detected boxes:
\begin{equation}
|x_2-x_1|>|y_2-y_1|.
\end{equation}
Horizontal reflection leaves this condition unchanged. Without the sign of $x_2-x_1$ and a consistent assignment of detections to the two named objects, the rule cannot tell ``left of'' from ``right of.'' The controlled spatial reversals test what this means for rendered videos.

\section{Detailed Baselines and Intervention Specifications}
\label{app:baselines}

\begin{figure}[ht]
\centering
\includegraphics[width=\textwidth,height=0.80\textheight,keepaspectratio]{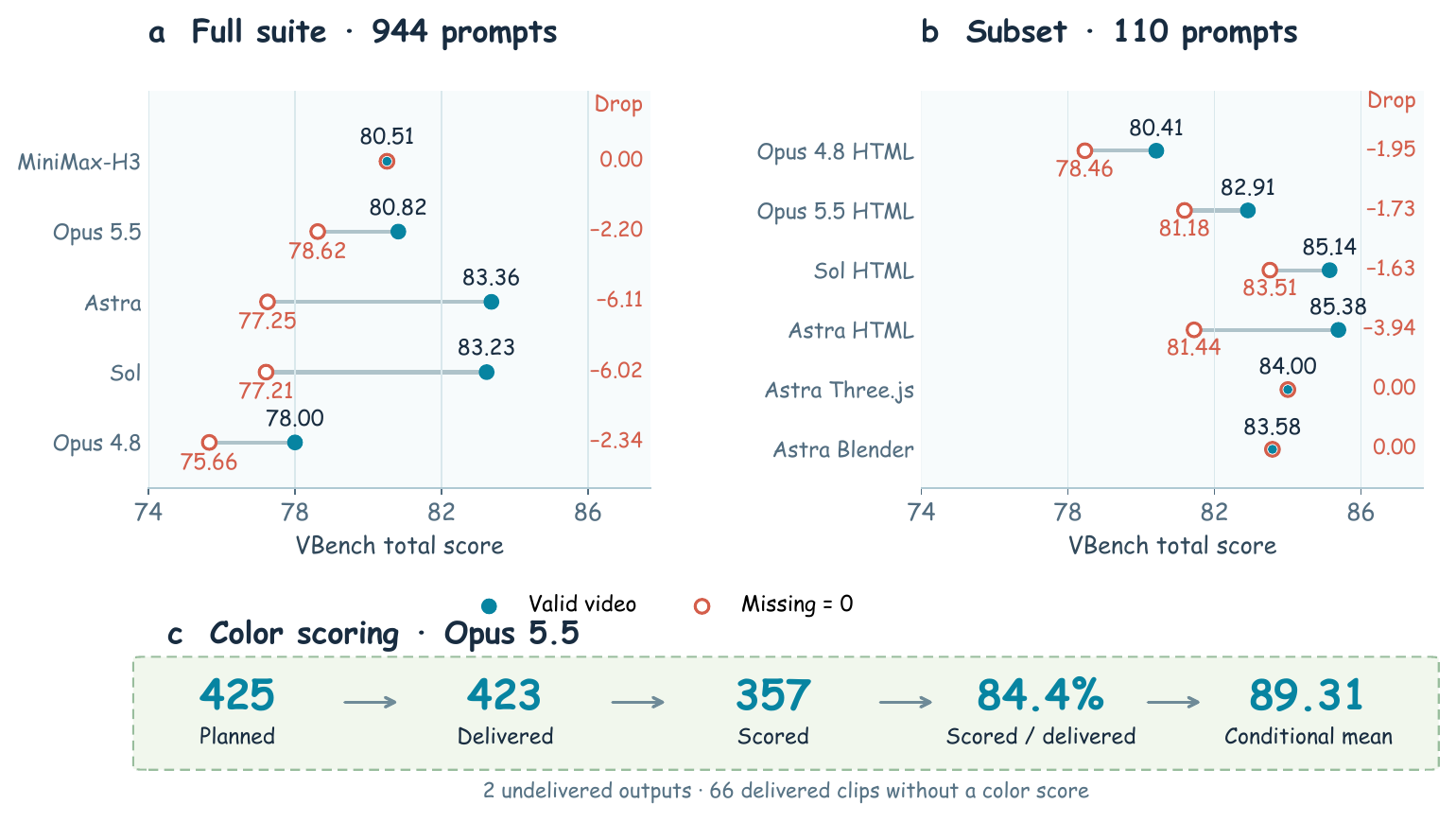}
\caption{\textbf{Delivery failures change rankings, while detector omissions reduce scoring coverage.} Filled blue dots show totals over delivered videos, and open coral dots give zero to undelivered videos. (a) The full suite includes four coding models and MiniMax-H3. (b) The 110-prompt subset compares direct HTML models and Astra's Three.js and Blender agents. Drops use the same outputs under both conventions. (c) Opus 5.5 color coverage: 425 planned, 423 delivered, and 357 scored; 66 delivered clips are unscored and two outputs are missing.}
\label{fig:scores}
\end{figure}

Figure~\ref{fig:scores} shows the full-suite and 110-prompt subset totals under both averaging conventions, next to the color-scoring coverage. Table~\ref{tab:controls} describes the six program edits used in Section~\ref{sec:controlled}.

Execution feedback improves delivery on the 110-prompt subset. Multi-turn coding agents deliver all 750 assigned outputs and raise GPT-6 Astra's total with missing videos counted as zero from 81.44 for direct HTML to 84.00 with Three.js and 83.58 with Blender, at $3\times$ to $30\times$ the latency (Figures~\ref{fig:scores} and~\ref{fig:exp-methods}). The comparison separates the quality of delivered videos from the chance of delivering one. The 983 full-suite HTML failures are broken down in Appendix~\ref{app:rendering}.

\begin{table}[ht]
\centering\small
\setlength{\tabcolsep}{5pt}
\caption{\textbf{Controlled pairs test behavior changes and color changes.} Each variant starts from a verified scene program, and we inspect the rendered pair to check the intended change and any unintended visible effect. Continuous-motion violations are tested alongside frozen-action controls.}
\label{tab:controls}
\begin{tabularx}{\columnwidth}{p{2.8cm}*{3}{>{\raggedright\arraybackslash}X}}
\toprule
Intervention & Changed factor & Protected factors & Diagnostic target\\
\midrule
Action freeze & Target motion stops & Person, props, scene & Simple missing-action control\\
Non-contact motion & Contact or outcome & Continuous motion and scene & Recognition of the target event\\
Spatial change & Required relation & Object identity and reference frame & Recognition of relation errors\\
Order change & Event sequence & Events and scene & Temporal relations\\
Camera interference & Camera motion & Target action remains absent & Reliance on generic movement\\
Hue change & Whole-frame hue & All task requirements & Preservation of correct decisions\\
\bottomrule
\end{tabularx}
\end{table}

\section{Detailed Experimental Visualizations}
\label{app:profiles_text}
\label{app:experimental-visualizations}
\begin{figure}[ht]
\centering
\includegraphics[width=0.70\textwidth,height=0.4\textheight,keepaspectratio]{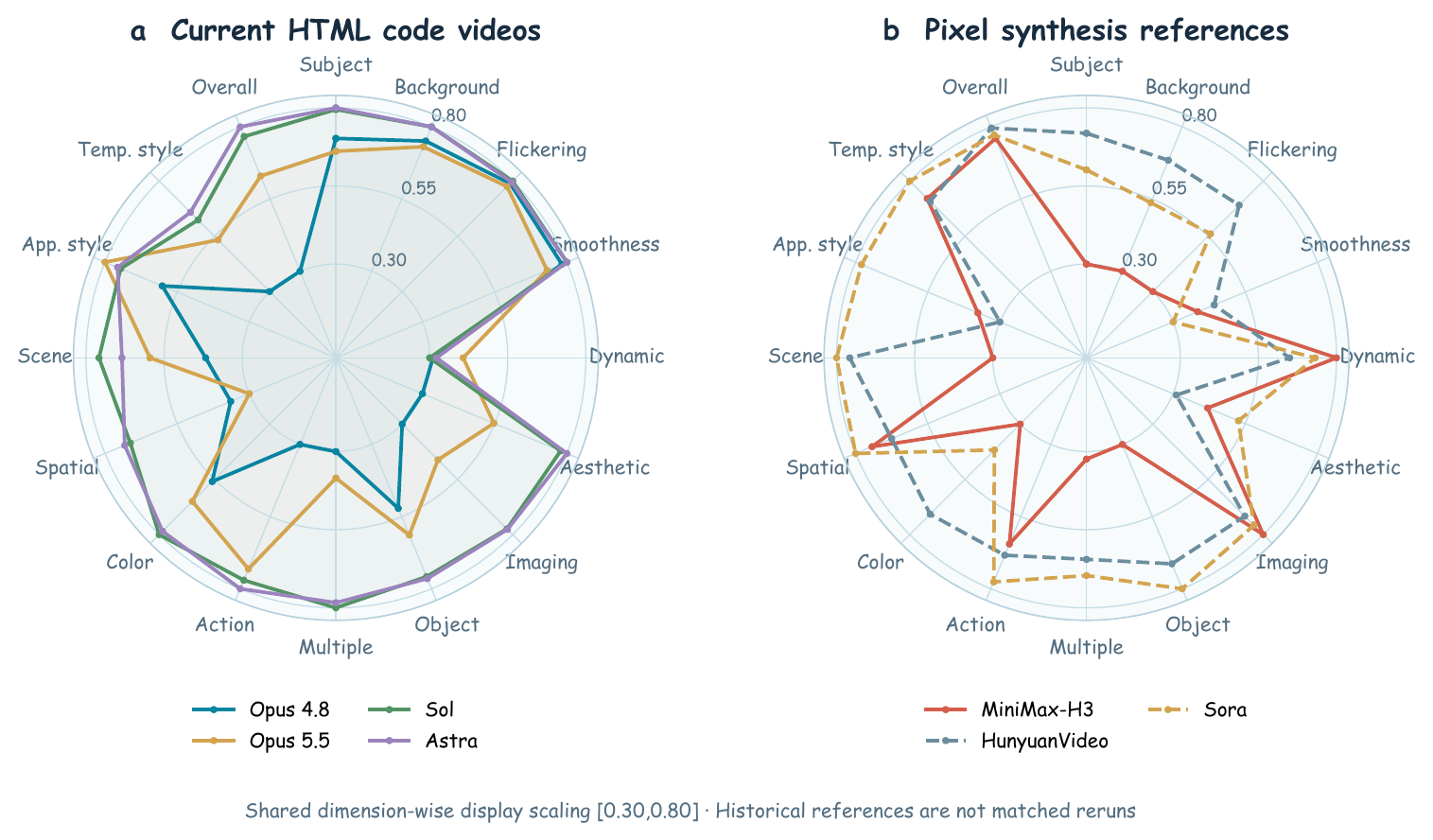}
\caption{\textbf{High rendering stability can coexist with limited motion.} (a) Four full-suite HTML models and (b) the matched MiniMax-H3 baseline plus historical HunyuanVideo and Sora references~\cite{kong2024hunyuanvideo,sora}. Each dimension uses a shared display scale of $[0.30,0.80]$, so these radii are not raw scores. Figure~\ref{fig:exp-dimensions} gives raw scores for our five evaluated systems. The historical references are not matched reruns.}
\label{fig:exp-radar}
\end{figure}

\begin{figure}[ht]
\centering
\includegraphics[width=0.8\textwidth,height=0.3\textheight,keepaspectratio]{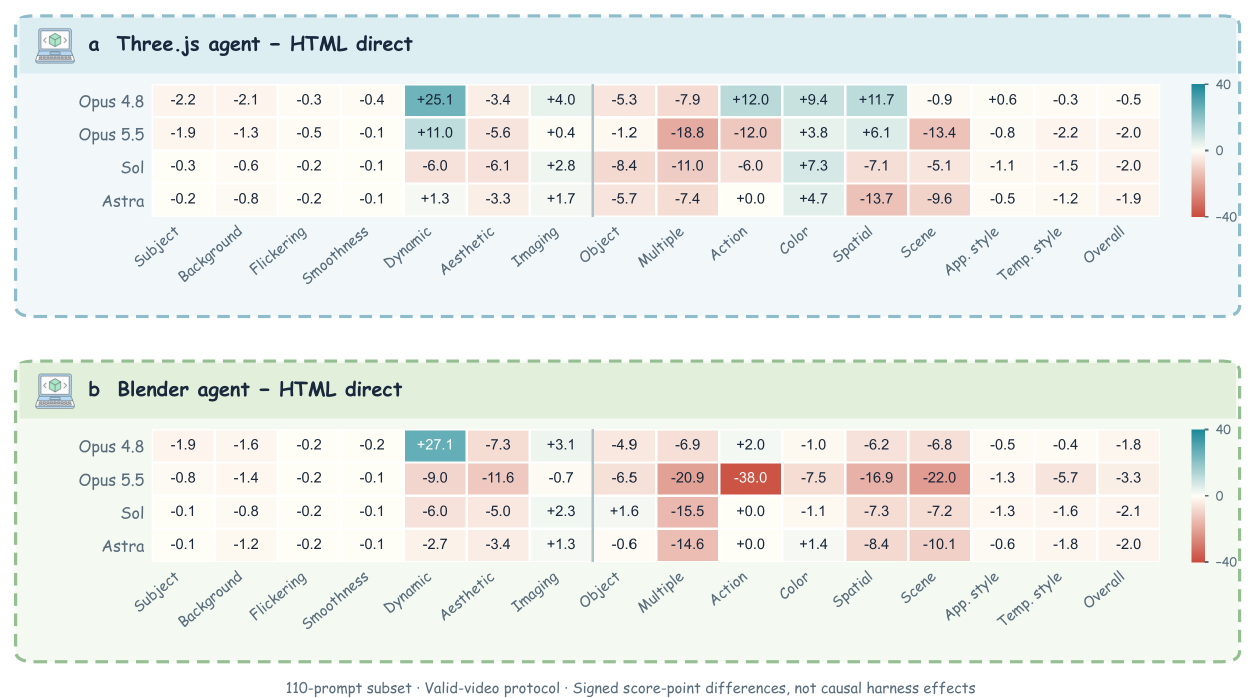}
\caption{\textbf{3D agent outputs receive higher dynamic-degree scores for Claude models alongside lower scene and multi-object scores.} Each cell is the difference in score over delivered videos (agent minus direct HTML, on a 0--100 scale) for (a) Three.js and (b) Blender agents on the 110-prompt subset across all 16 dimensions (blue positive, coral negative).}
\label{fig:exp-methods}
\end{figure}

\begin{figure}[ht]
\centering
\includegraphics[width=0.80\textwidth,height=0.3\textheight,keepaspectratio]{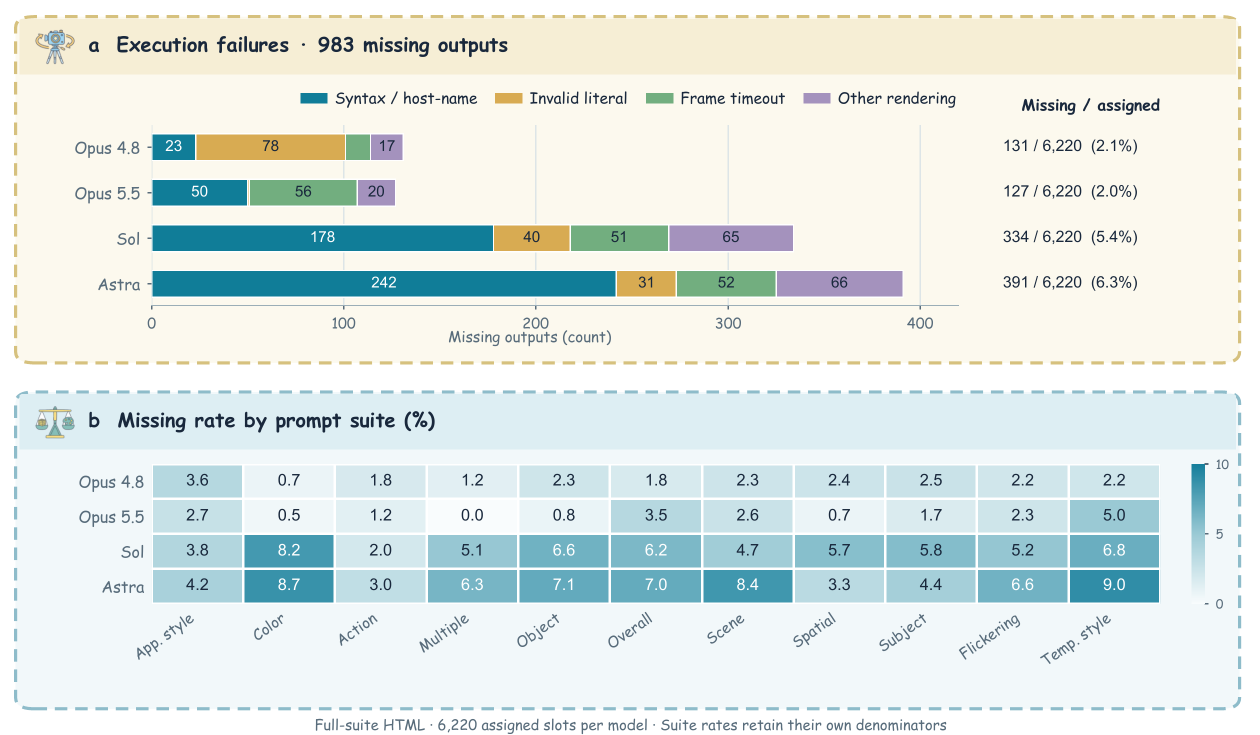}
\caption{\textbf{Half of full-suite HTML rendering failures involve syntax or identifier collisions.} (a) The 983 missing videos are grouped by model and recorded cause: syntax or host-name collisions (blue), invalid color literals (gold), frame timeouts (green), and other rendering errors (purple). Each model has 6,220 planned outputs. (b) Missing rates by prompt suite, each over its own assigned outputs.}
\label{fig:exp-execution}
\end{figure}

\begin{figure}[ht]
\centering
\includegraphics[width=\textwidth,height=0.39\textheight,keepaspectratio]{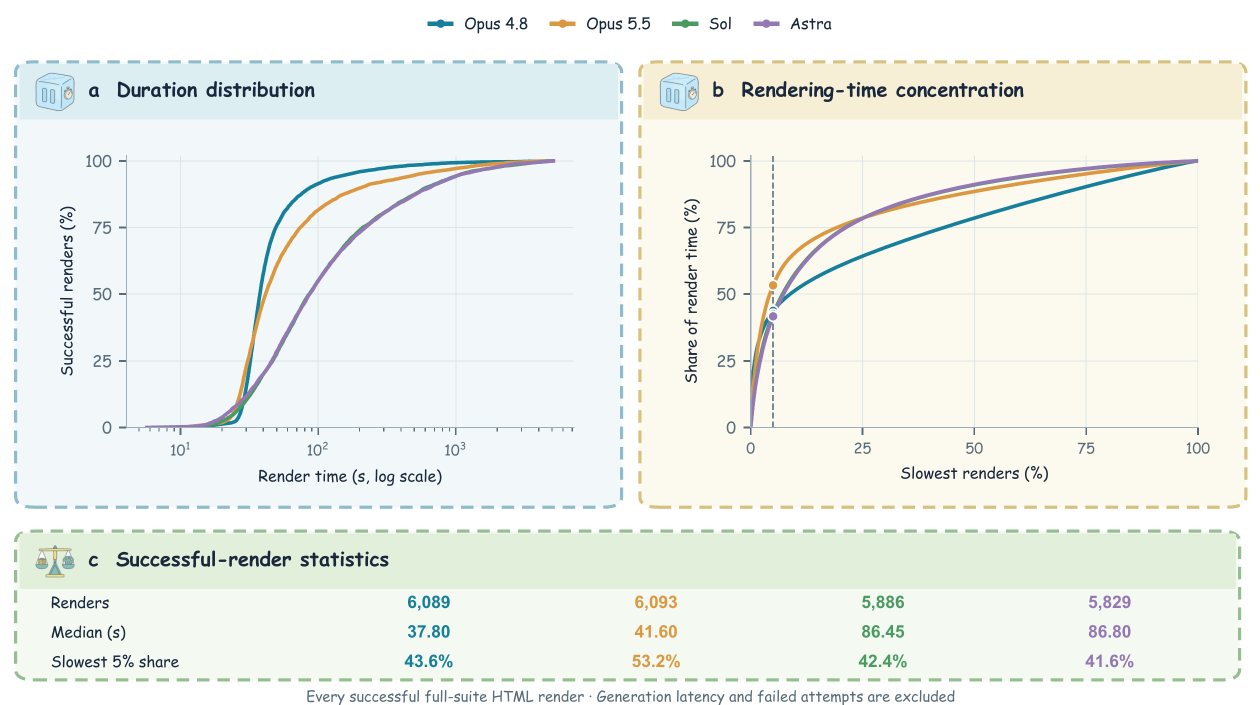}
\caption{\textbf{Full-suite HTML CPU rendering is heavy-tailed, with the slowest 5\% of scenes consuming 42\%--53\% of total runtime.} (a) Empirical duration CDF (logarithmic axis) and (b) complementary cumulative rendering-time concentration across all successful full-suite HTML renders for the four coding models.}
\label{fig:exp-timing}
\end{figure}

\begin{figure}[ht]
\centering
\includegraphics[width=\textwidth,height=0.80\textheight,keepaspectratio]{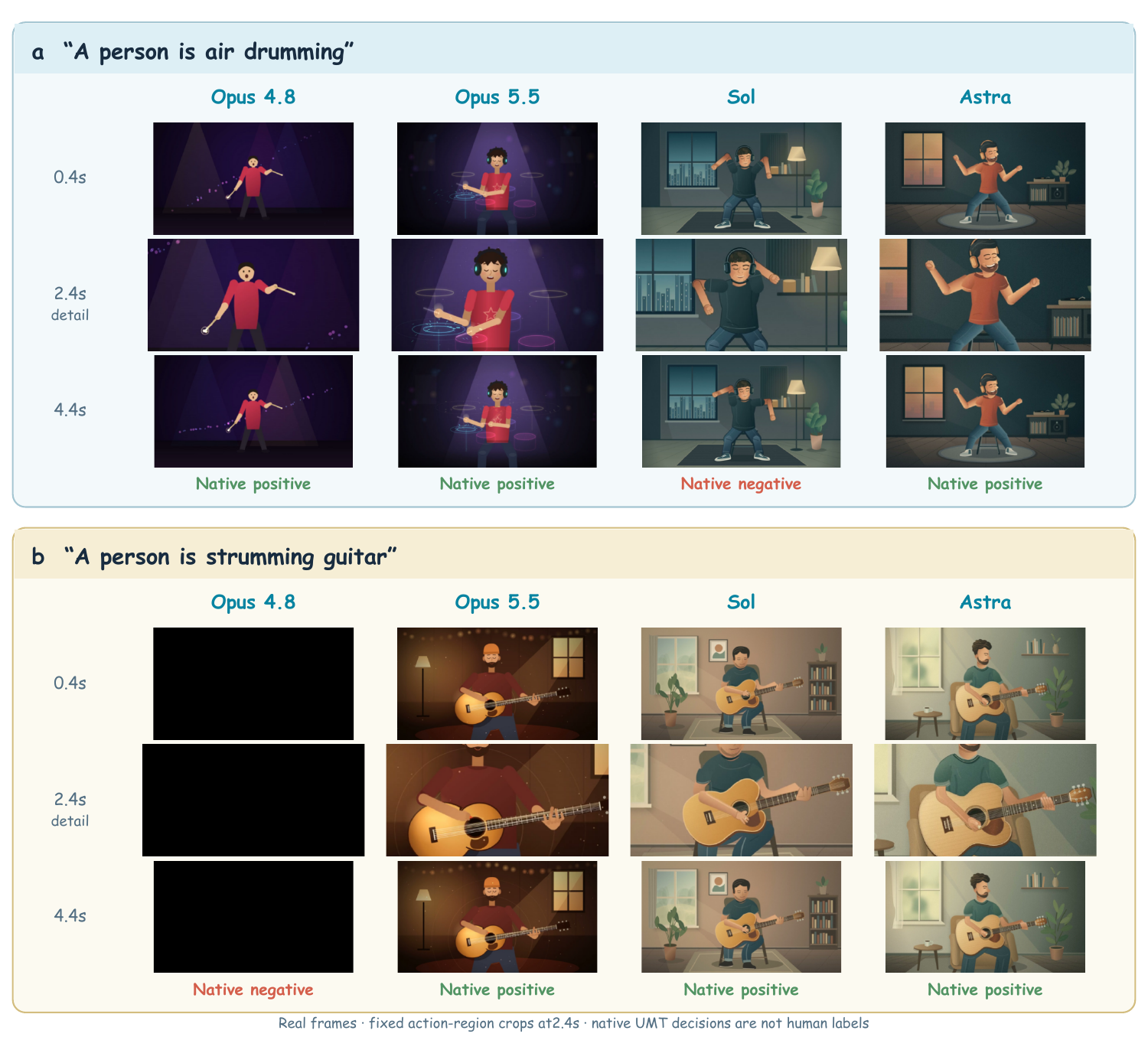}
\caption{\textbf{Native action decisions vary across renderings of the same request.} Eight full-suite HTML videos for ``air drumming'' and ``strumming guitar'' show full frames at 0.4 and 4.4 seconds, with a fixed action-region crop at 2.4 seconds. Green and coral labels denote positive and negative native UMT decisions, and they are not human completion labels. The Opus 4.8 guitar clip shows black frames at the three displayed times.}
\label{fig:exp-cases}
\end{figure}

Figures~\ref{fig:exp-radar} to~\ref{fig:exp-cases} show per-dimension profiles, exact scores, production-route differences, execution failures, rendering runtimes, and qualitative comparisons for the full suite and the 110-prompt subset.

\subsection{Dimension-Level Profiles}
\begin{figure}[ht]
\centering
\includegraphics[width=0.8\textwidth,height=0.3\textheight,keepaspectratio]{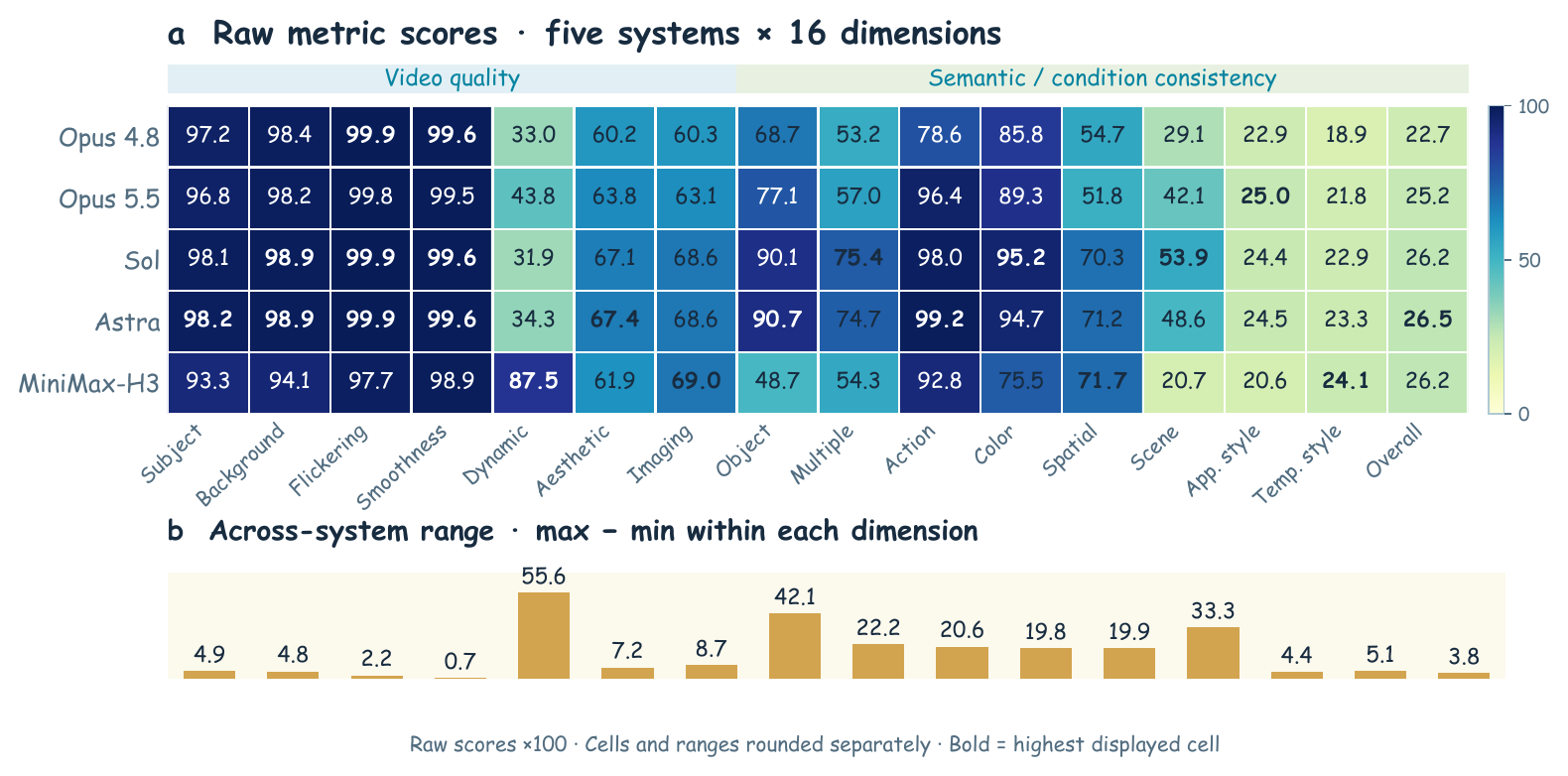}
\caption{\textbf{Motion and semantic dimensions separate code rendering from MiniMax-H3.} (a) Raw values ($\times100$) for four full-suite HTML models and MiniMax-H3, with the highest displayed value per dimension in bold. Blue and green headers separate video quality and semantic consistency. (b) The maximum-minus-minimum range across these five systems. Cells and ranges are rounded separately to one decimal.}
\label{fig:exp-dimensions}
\end{figure}

Code rendering is stable but moves little (Figures~\ref{fig:exp-radar} and~\ref{fig:exp-dimensions}). Browser rendering has no diffusion noise or camera shake, so all four HTML models reach 99.81 to 99.92 on temporal flickering and 99.50 to 99.62 on motion smoothness, ahead of MiniMax-H3 on flickering (97.72), smoothness (98.89), subject consistency (96.81--98.16 vs.\ 93.30), and background consistency (98.21--98.87 vs.\ 94.07). On dynamic degree, however, HTML models score only 31.86 to 43.79, less than half of MiniMax-H3 (87.50). They match MiniMax-H3 on the aggregate Quality score (83.26--84.43 vs.\ 84.57) and lead on the Semantic score (71.02--79.45 vs.\ 64.24). Perceptual quality is mixed: HTML aesthetic quality spans 60.20--67.40, and three of the four HTML models exceed MiniMax-H3 (61.92); on matched prompts, GPT-6 Astra and GPT-6.1 Sol score higher than MiniMax-H3 on 72\% and 74\% of clips. Imaging quality (60.27--68.61) stays below MiniMax-H3's 69.02 for all four. These aesthetic means cover the full dimension and not the matched human-audit clips.

\subsection{Raw Scores and Production-Route Differences}

\subsection{Execution Failures and Rendering Runtime}

\clearpage
\subsection{Qualitative Comparisons}

\end{document}